\pdfoutput=1

\documentclass[11pt]{article}

\usepackage[blocks]{authblk}
\usepackage[final]{acl}

\usepackage{times}
\usepackage{latexsym}

\usepackage[T1]{fontenc}

\usepackage[utf8]{inputenc}

\usepackage{microtype}

\usepackage{inconsolata}

\usepackage{graphicx}

\usepackage{amsmath}
\usepackage{amsfonts}
\usepackage{bm}
\usepackage{multirow}
\usepackage{tabularx}
\usepackage{array}
\usepackage{color}
\usepackage{enumitem}
\usepackage{makecell}
\usepackage{xcolor}
\usepackage{colortbl}
\usepackage{booktabs}
\usepackage{algorithm}
\usepackage{algpseudocode}
\usepackage{subcaption}
\usepackage{graphics}
\usepackage{booktabs}
\usepackage{relsize}
\usepackage{xcomment}

\newcommand{\rev}[1]{{\color{black}{#1}}}

\newcommand{\eg}{\textit{e.g.},}

\definecolor{my_blue}{RGB}{230, 230, 250}
\definecolor{my_green}{RGB}{230, 250, 230}

\newcolumntype{C}{>{\centering\arraybackslash}X}
\newcolumntype{Y}{>{\centering\arraybackslash}p{1.75cm}}
\newcolumntype{L}{>{\raggedright\arraybackslash}l}
\newcolumntype{B}{>{\columncolor{my_blue}}C}

\title{By My Eyes: Grounding Multimodal Large Language Models\\with Sensor Data via Visual Prompting}

\makeatletter 
\renewcommand\AB@affilsepx{   \protect\Affilfont}
\makeatother

\author[1]{\textbf{Hyungjun Yoon}}
\author[1]{\textbf{Biniyam Aschalew Tolera}}
\author[2]{\authorcr \textbf{Taesik Gong}}
\author[1]{\textbf{Kimin Lee}}
\author[1]{\textbf{Sung-Ju Lee}}
\affil[1]{KAIST}
\affil[2]{UNIST}
\affil[ ]{\authorcr \small{\texttt{\{hyungjun.yoon, binasc, kiminlee, profsj\}@kaist.ac.kr, taesik.gong@unist.ac.kr}}}

\begin{document}
\maketitle

\begin{abstract}

Large language models (LLMs) have demonstrated exceptional abilities across various domains. However, utilizing LLMs for ubiquitous sensing applications remains challenging as existing text-prompt methods show significant performance degradation when handling long sensor data sequences. We propose a visual prompting approach for sensor data using multimodal LLMs (MLLMs). We design a visual prompt that directs MLLMs to utilize visualized sensor data alongside the target sensory task descriptions. Additionally, we introduce a visualization generator that automates the creation of optimal visualizations tailored to a given sensory task, eliminating the need for prior task-specific knowledge. We evaluated our approach on nine sensory tasks involving four sensing modalities, achieving an average of 10\% higher accuracy than text-based prompts and reducing token costs by 15.8$\times$. Our findings highlight the effectiveness and cost-efficiency of visual prompts with MLLMs for various sensory tasks. \rev{The source code is available at \url{https://github.com/diamond264/ByMyEyes}}.


\end{abstract}
\section{Introduction}



Large language models (LLMs) have shown remarkable performance in tasks across diverse domains, including science, mathematics, medicine, and psychology~\cite{bubeck2023sparks}. The recent advent of multimodal LLMs (MLLMs), \eg{} GPT-4o~\cite{openai2024gpt4o}, has further expanded their capabilities to images and audio inputs, broadening their use in fields such as industry and medical imaging~\cite{yang2023dawn}. 

\rev{Meanwhile, sensor data---including measurements from smartphones, wearables, IoT~\cite{dian2020wearables}, and medical equipment~\cite{pantelopoulos2009survey}---holds potential for ubiquitous applications when effectively integrated with MLLMs. Sensory tasks involve extensive and significant applications, ranging from authentication~\cite{abuhamad2020sensor} and healthcare~\cite{wang2019survey} to agriculture~\cite{sishodia2020applications} and environmental monitoring~\cite{feng2019review}. 
However, MLLMs remain underutilized.} 
The diversity of sensors~\cite{wang2019survey} and the heterogeneity among them~\cite{stisen2015smart} hinder the implementation of a foundational model that generalizes across various sensing tasks. The expensive data collection~\cite{vijayan2021review} often results in insufficient training data, further complicating the development of such capability.

\begin{figure}[t]
    \includegraphics[width=\columnwidth]{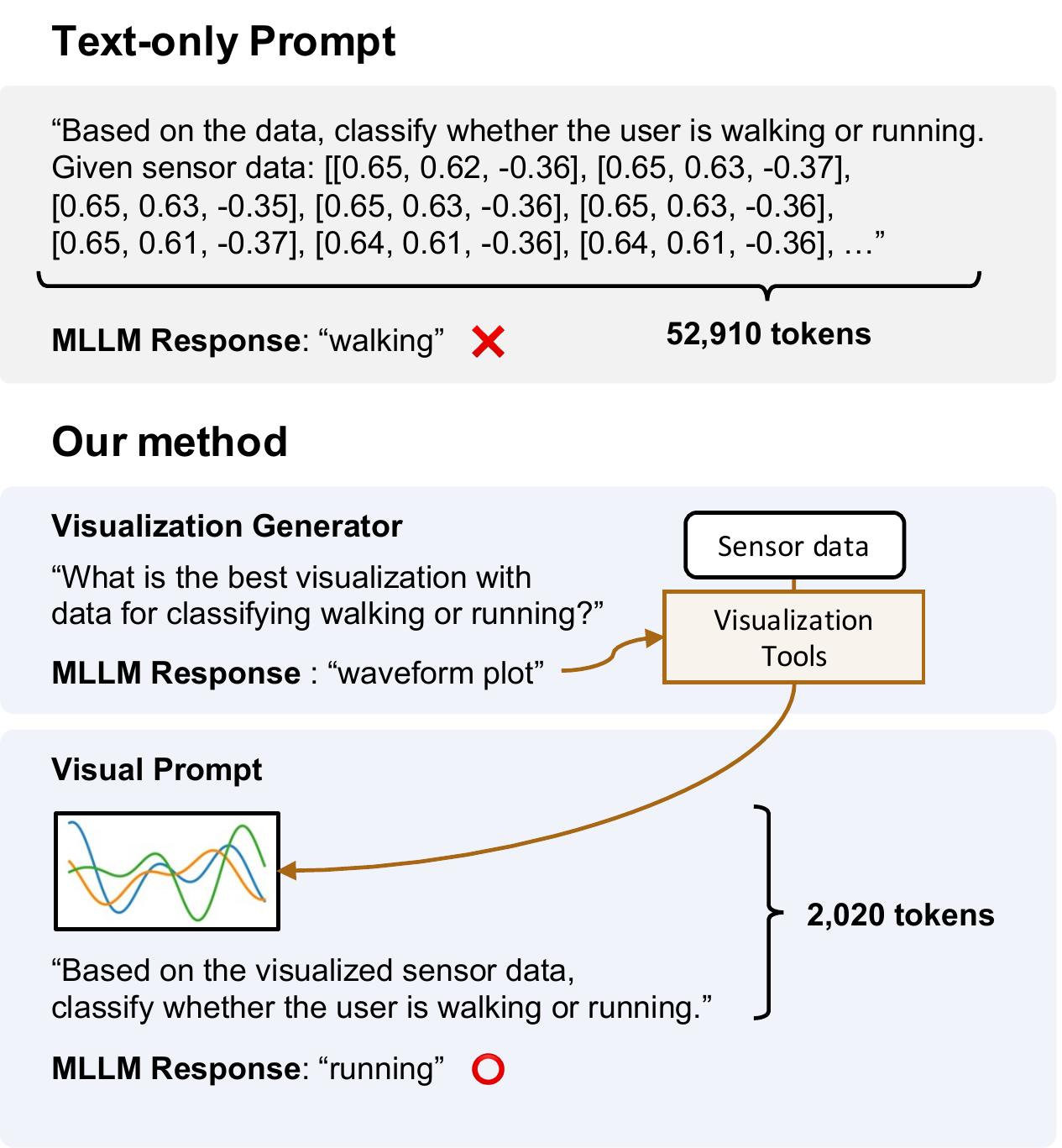}
    \caption{An example of solving a sensory task using an MLLM with visual prompts. The visualization generator generates an appropriate visualization for the given sensor data, and the visualized data is provided as an image to the MLLM for solving the task.}
    \label{fig:intro_example}
    \vspace{-10pt}
\end{figure}

Recent studies explored leveraging pre-trained LLMs to solve general sensory tasks~\cite{yu2023zero, liu2023large, kim2024health}. 
One approach extracts task-specific features from sensor data and composes them as prompts~\cite{yu2023zero}. However, designing such prompts requires specific domain knowledge. Alternatively, incorporating raw sensor data as text prompts~\cite{kim2024health, liu2023large} has been a widely used method to handle sensory data with LLMs as a more generalizable solution. Yet, we empirically found that providing raw sensor data with text prompts shows poor performance in real-world sensory tasks with long-sequence inputs and incurs high costs due to an extensive number of tokens.

To address these challenges, we propose providing visualized sensor data as images to MLLMs that support visual inputs. Leveraging MLLMs' growing ability to interpret visual aids~\cite{yang2023dawn}, we explore their effectiveness in analyzing plots generated from sensor data. We designed a visual prompt comprising visualized sensor data and task-specific instructions to solve sensory tasks. In addition, we present a visualization generator that enables MLLMs to independently generate optimal visualizations using tools available in public libraries. This generator filters potential visualization methods based on the task description and assesses the resulting visualizations of each method to determine the best visualization. Figure~\ref{fig:intro_example} compares the existing text-only prompts with our method for sensory tasks.


Evaluations on nine sensory tasks involving four different modalities showed that visual prompts generated from the visualization generator significantly improved performance by an average of 10\% while reducing token costs by 15.8$\times$ compared with the existing baseline. Our findings highlight the effectiveness and efficiency of visualized sensor data with MLLMs in various applications.

We summarize our contributions as follows:

\begin{itemize}
\itemsep 0em
    \item We propose to ground MLLMs with sensor data by providing visualized sensor data as images, achieving improved performance at reduced costs than the text-based baseline.
    \item We present a visualization generator that automatically generates suitable visualizations for various sensory tasks using public libraries.
    \item We conduct experiments on nine different sensory tasks across four modalities, demonstrating the broad applicability of our approach.
\end{itemize}
\vspace{1pt}

\section{Related Work}

\textbf{LLMs with sensor data.} Sensory tasks involve sequences of numbers indicating values over time. Initial research for handling sequential data focused on time-series forecasting~\cite{zhang2024large}. Converting time-series data into text prompts for forecasting has been proposed in PromptCast~\cite{xue2023promptcast} and LLMTime~\cite{gruver2024large}. Other studies~\cite{zhou2023one, jin2023time} used specialized encoders to create embeddings compatible with pre-trained LLMs.

Beyond forecasting, LLMs have been explored in healthcare for their ability to answer questions using physiological sensor data~\cite{liu2023large}. For example, LLMs have been used for ECG diagnosis~\cite{yu2023zero} by integrating ECG-specific features and knowledge from ECG databases. Penetrative AI~\cite{xu2024penetrative} and Health-LLM~\cite{kim2024health} have used raw sensor data in text prompts to solve health problems without task-specific processing. Our study examines whether existing methods can generalize to broader sensing tasks with high-frequency, long-duration data. Building upon these works, we propose visualizing sensor data for MLLMs to improve their performance and cost efficiency.

\textbf{Multimodal large language models (MLLMs).} Advancements in MLLMs~\cite{zhang2024mm} have equipped popular models such as ChatGPT~\cite{openai2022chatgpt} with vision capabilities~\cite{openai2024gpt4o}. Recent studies explored the in-context learning~\cite{brown2020language} abilities of MLLMs, showing that they can understand images with the interleaved text and few-shot examples~\cite{tsimpoukelli2021multimodal, alayrac2022flamingo}. This capability has been applied in medical diagnostics, including analyzing radiology and brain images with accompanying text instructions~\cite{wu2023can}. Our work explores using MLLMs to analyze visualized sensor data for broader applications.

\textbf{Using tools with LLMs.} Recent research has shown that augmenting LLMs with external tools can extend their capabilities. Toolformer~\cite{schick2024toolformer} enables LLMs to access public APIs and search engines, while Visual Programming~\cite{gupta2023visual} uses LLMs to generate and execute codes. HuggingGPT~\cite{shen2024hugginggpt} and Chameleon~\cite{lu2024chameleon} integrated multiple expert models to enhance functionalities. 
\rev{Recently, Data Interpreter~\cite{hong2024data} enabled LLMs to analyze data and build task-specific models for data interpretation. Building on them, our work leverages MLLMs to utilize sensor data visualization tools. Importantly, unlike existing approaches that rely on external tools as the primary task solvers, we propose positioning MLLMs as the main solvers for sensory tasks, leveraging their in-context learning abilities. Our approach aims to eliminate the need for additional data collection and training, thereby addressing the scarcity of public resources for sensory tasks. Furthermore, we introduce a design in which MLLMs perform demonstration-based assessments to evaluate their task-solving effectiveness, ensuring optimal visualization for specific tasks.}

\section{Limitations of Representing Sensor Data as Text-based Prompts}
\label{sec:motivation}


Existing approaches for grounding language models with sensor data primarily rely on text-based prompts~\cite{liu2023large, jin2023large, zhang2024large, yu2023zero}. 
One approach uses prompts with specialized features extracted from sensor data for specific tasks, such as R-R intervals for ECG-based applications~\cite{yu2023zero}. While this approach effectively handles known sensory tasks, feature-based prompts often require domain knowledge, which is not generalizable for non-expert users. Instead, a more common approach~\cite{kim2024health, xu2024penetrative, liu2023large} incorporates raw sensor data sequences directly into prompts without data processing. However, most studies focus on short sequences (e.g., fewer than 100 elements)~\cite{kim2024health} and simple tasks (e.g., binary classification)~\cite{liu2023large}.

Real-world sensor data often entail long numeric sequences with high sampling rates and long durations. For example, arrhythmia detection~\cite{wagner2020ptb} requires ECG data sampled at 100Hz over 10 seconds, resulting in 1,000 elements. This section investigates the limitations of using text-based prompts to represent such complicated sensor data in language models. We focus on the capability to interpret sensor data and the token consumption costs associated with long numeric sequences. 

\textbf{Language models struggle to interpret long numeric text sequences.} Language models interpret simple numeric sequences by performing arithmetic operations~\cite{achiam2023gpt} and understanding sequential data~\cite{gruver2024large, mirchandani2023large}. However, we empirically revealed that their performance declines significantly with longer sequences inside the prompt, such as those exceeding 100 numbers, common in sensor data.

We conducted experiments with two specific tasks: \textit{mean prediction} to evaluate arithmetic capabilities~\cite{pirttikangas2006feature} and \textit{wave classification} to assess pattern recognition~\cite{liu2016complex} in sequences. The defined tasks represented the basic functionalities for sensor data interpretation, serving as typical feature extraction methods. Using randomly generated sine and sawtooth waves with varying lengths, we asked a language model, GPT-4o~\cite{openai2024gpt4o}, to calculate mean values and classify wave types using one-shot examples for each task. Each task was repeated 30 times to ensure robustness.

\begin{figure}[t]
    \includegraphics[width=\columnwidth]{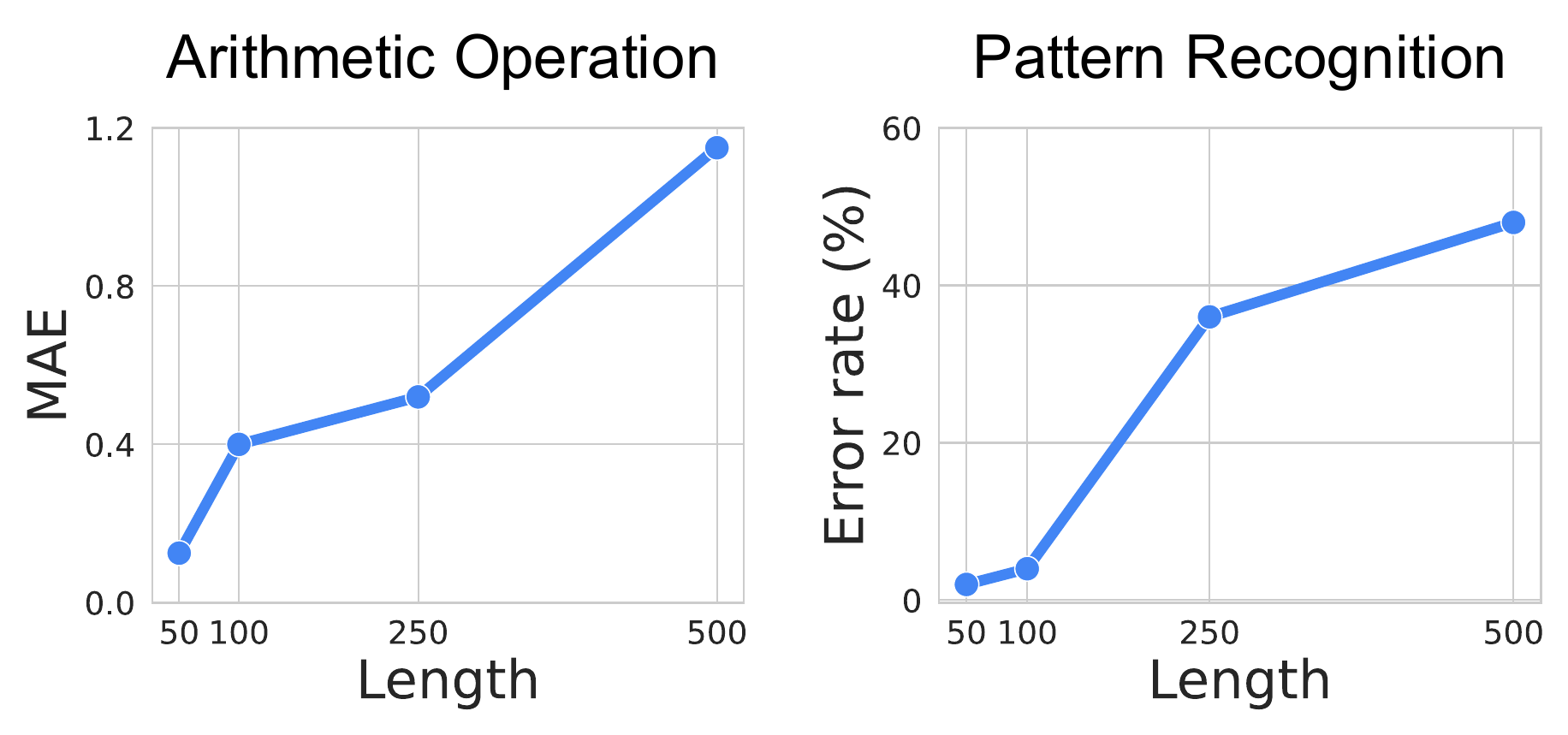}
    \caption{Performance of GPT-4o on arithmetic operation (mean prediction) and pattern recognition (sine and sawtooth wave classification) tasks for varying lengths.}
    \label{fig:motivation_study}
\end{figure}

Figure~\ref{fig:motivation_study} shows the results. In arithmetic operations, error rates consistently increased with the length. In pattern recognition, performance declines significantly for sequences longer than 100 elements, approaching the performance of a random classifier at 500 elements. 
While recent models such as GPT-4o, with its 128K context window, support long input lengths, our results indicate that interpreting sensor data with long numeric sequences still remains challenging.

\begin{figure*}[t]
    \centering
    \includegraphics[height=6cm]{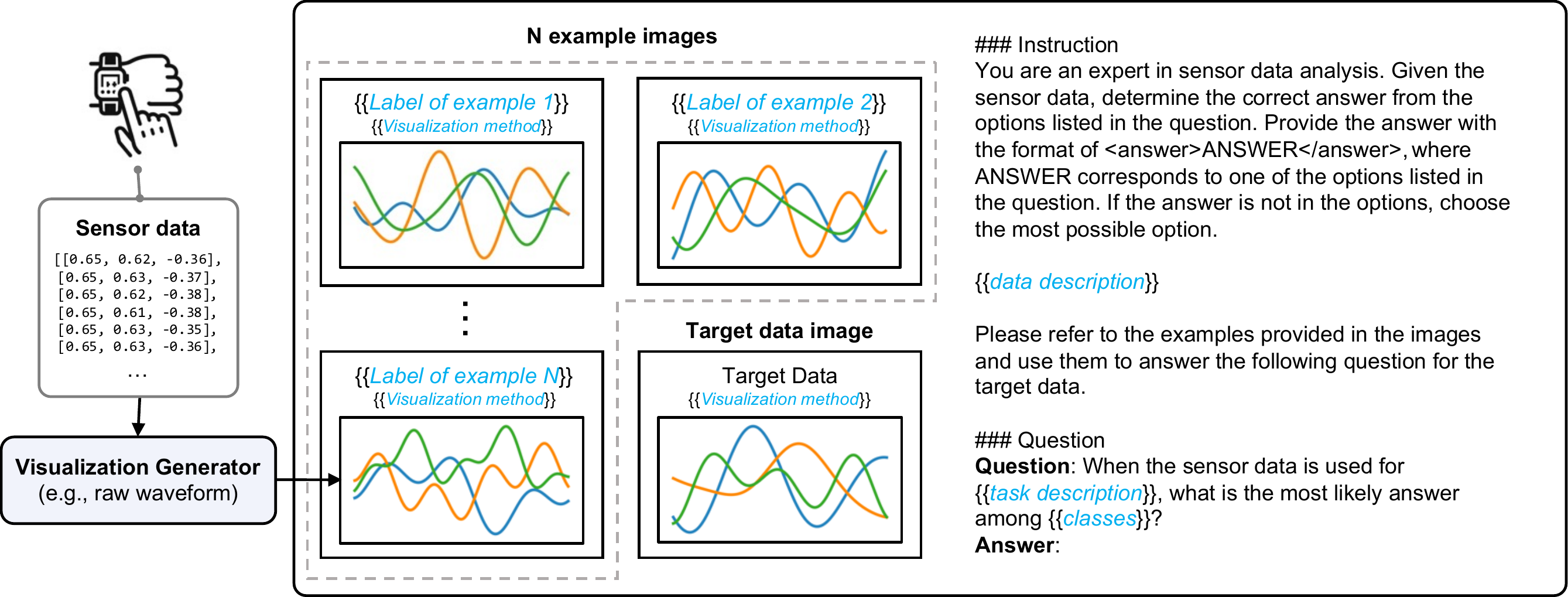}
    \caption{Overview of our visual prompt. 
    Sensor data are transformed into annotated images with labels and visualization methods.
    Additionally, instructions are provided to the MLLM, detailing the task and relevant data descriptions. These instructions guide the MLLM on effectively utilizing the provided images to solve the task.}
    \label{fig:method_visual_prompt_template}
\end{figure*}

\textbf{Sensor data in text is costly.} 
The computational and financial burdens for API users of language models scale with the number of tokens in the prompt.
Representing sensor data in textual format leads to extensive token usage, thereby increasing costs. 
For instance, performing passive sensing to track activities with smartphone accelerometers~\cite{stisen2015smart} uses sensor data sampled at 100~Hz. Collecting this data over a minute results in a prompt of 18K numbers, translating to 90K tokens. This leads to a huge cost of \$450 per hour when using GPT-4o API to classify six activities with one example for each. Higher sampling rates or longer durations further increase the costs, making such applications infeasible.

\textbf{Transition to visual prompts.} 
Language models such as ChatGPT (GPT-4o~\cite{openai2024gpt4o}) and Gemini~\cite{deepmind2024gemini} have expanded capabilities to include multimodal inputs (e.g., vision and audio). Recent Multimodal Large Language Models~(MLLMs) demonstrate an increasing ability to identify patterns and interpret visual data~\cite{achiam2023gpt}. This opens new opportunities for sensory tasks, as sensor data are often visualized for analysis. Visualizations make complex data more interpretable and condense long data sequences into a \textit{single} image, significantly reducing token costs. Building on this capability, we exploit visualized sensor data instead of text-based prompts.

\section{Method}

We introduce our method for handling sensory tasks by providing sensor data as image inputs to MLLMs.  Section~\ref{method:visual_prompt_design} overviews our prompt design strategy. 
Section~\ref{method:visualization_planner} introduces our visualization generator, which automatically generates suitable visualizations for heterogeneous sensor data.

\subsection{Visual Prompt Design}
\label{method:visual_prompt_design}

To leverage MLLMs for sensory tasks, we propose a {\em visual prompt}, as illustrated in Figure~\ref{fig:method_visual_prompt_template}.
The key idea is to transform numeric sequences of sensor data into visual plots using various methods, such as raw waveforms and spectrograms. 
Detailed information about these visualization methods is in Section~\ref{method:visualization_planner}.
For few-shot examples, each plot includes a label as a title above it (i.e., \textit{\{\{Label of example X\}\}}). 
For unlabeled target data used in queries, the title is simply stated as ``target data''.
We provide textual instructions to clarify the data collection process and the task's objectives.
These instructions ensure that MLLMs can effectively interpret and utilize the visualized sensor data.

\begin{figure*}[t]
    \centering
    \includegraphics[width=2\columnwidth]{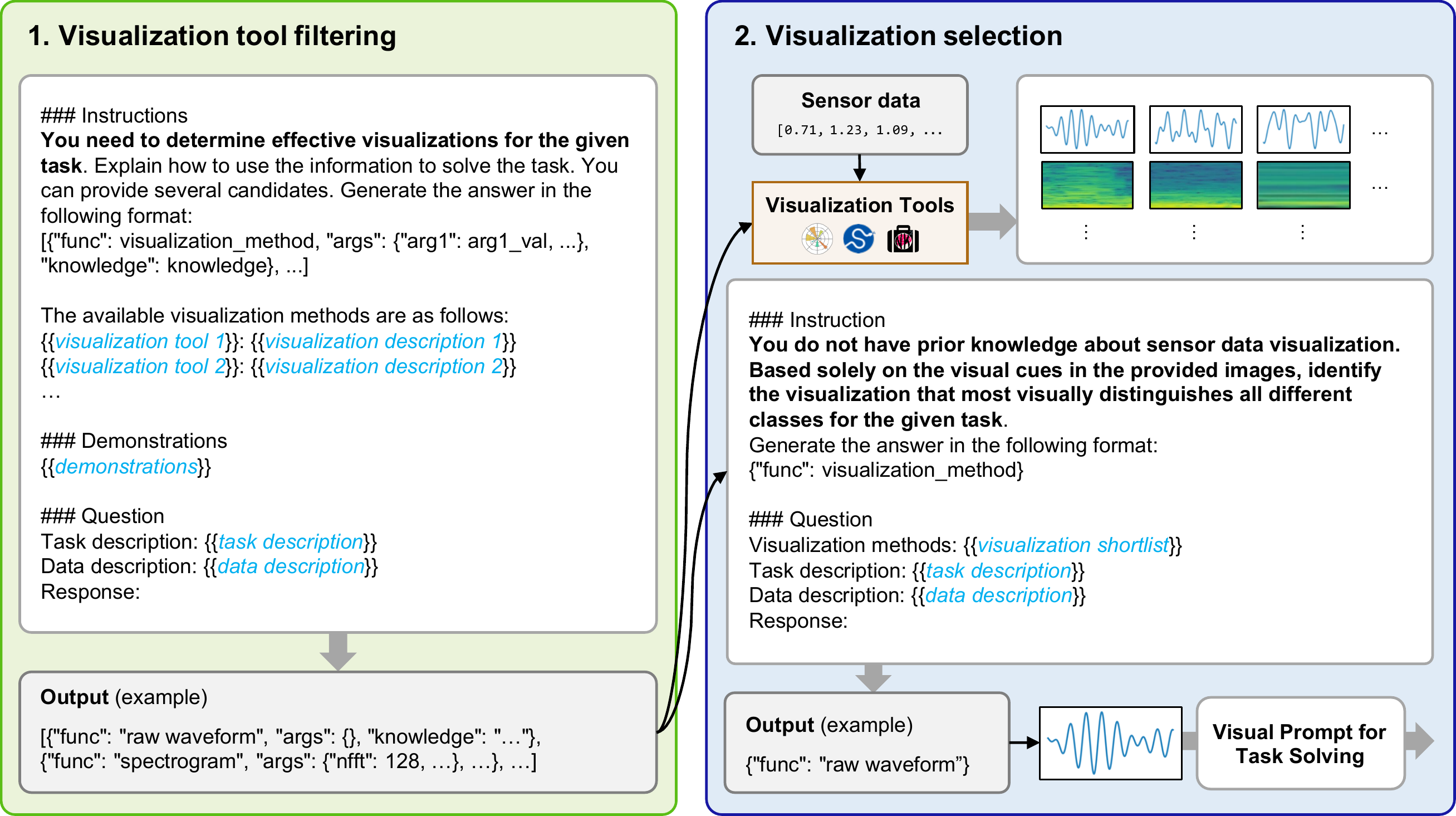}
    \caption{Overview of our visualization generator. First, visualization tool filtering generates a filtered list of visualization tools from public libraries based on the task and data descriptions. Next, visualization selection generates and selects the most effective visualization by asking MLLMs to observe visualized sensor data prepared for the task using all the filtered visualization methods.}
    \label{fig:method_visualization_planner}
\end{figure*}

\subsection{Visualization Generator} \label{method:visualization_planner}

In our proposed visual prompt, the choice of visualization method is crucial, as it significantly influences the MLLM's ability to comprehend the sensor data.
For example, raw waveform plots are ideal for tasks involving amplitude pattern recognition over time, while spectrograms~\cite{ito2018application} are suitable for tasks relying on frequency features.
We introduce a \textit{visualization generator} that automatically chooses the most suitable visualization tool from available public libraries, enabling non-expert users to effectively utilize visual prompts.
This generator operates in two main phases: (i)~visualization tool filtering and (ii)~visualization selection (see Figure~\ref{fig:method_visualization_planner}).

\noindent\textbf{Visualization tool filtering.} Public libraries offer a vast array of sensor data visualizations. However, trying each out to identify the optimal visualization is computationally expensive. To minimize the cost, we employ a filtering approach. By providing available visualization tools, descriptions of the task, and data collection, we ask MLLMs to select a list of visualization methods useful for the target task.

As shown in Figure~\ref{fig:method_visualization_planner} (green box), we provide a full list of available visualization methods found in public libraries (\eg{} Matplotlib~\cite{hunter2007matplotlib}, Scipy~\cite{virtanen2020scipy}, and Neurokit2~\cite{makowski2021neurokit2}) along with task and data collection descriptions to MLLMs. We also leverage the in-context learning ability of MLLMs to enhance response quality by providing demonstrations of optimal visualizations chosen for different tasks. The MLLM is instructed to output a list in JSON format, which is suitable for automated parsing at a later stage. Appendix~\ref{appendix:visualization_tools} shows the full list of available visualization tools and demonstrations.



\noindent\textbf{Visualization selection} Sensor data exhibit variations by instance due to user-specific behaviors, environmental factors, or device settings~\cite{stisen2015smart}, which cannot be fully captured in task and data descriptions. This variability limits the reliability of selecting visualizations based solely on the descriptions. To address this, we visualize the sensor data using all filtered visualization tools and ask the MLLM to select the one that provides the best visual information for the task.

The blue box in Figure~\ref{fig:method_visualization_planner} illustrates this procedure. First, different visualizations are generated using the filtered tools. With the images, we instruct the MLLM to select the best visualization by providing a textual prompt, including the visualization methods, task, and data details. We found that MLLM often makes incorrect decisions by prioritizing the task description over the visual aids. To prioritize visual efficacy, we explicitly instruct the MLLM to avoid relying on prior knowledge about sensor data and focus on the provided images. Finally, our automated framework conveys the selected visualization to the visual prompt for task solving.
\section{Experiments}

We evaluate the applicability of our approach with MLLMs by conducting experiments on a range of sensory tasks. 

\begin{table*}[t]
\centering
\caption{Comparison of the text prompts and visual prompts for solving sensory tasks using GPT-4o. The highest accuracy values are highlighted in bold. The visual prompts (multi-shot) utilize the maximum number of examples by matching the token size of the 1-shot text prompts.}
\label{tab:experiments_main_eval}
\small
{\renewcommand{\arraystretch}{1.1}
\begin{tabularx}{\textwidth}{LCCCCCCCCCC}
\Xhline{2\arrayrulewidth}
\addlinespace[0.1cm] 

 & \multicolumn{3}{c}{Accelerometer} & \multicolumn{4}{c}{ECG} & EMG & Resp & \\
\cmidrule(lr){2-4} \cmidrule(lr){5-8} \cmidrule(lr){9-9} \cmidrule(lr){10-10}
\multirow{2}{*}{Method} & \multirow{2}{*}{HHAR} & \multirow{2}{*}{\shortstack{UTD-\\MHAD}} & \multirow{2}{*}{Swim} & \multirow{2}{*}{\shortstack{PTB-XL\\(CD)}} & \multirow{2}{*}{\shortstack{PTB-XL\\(MI)}} & \multirow{2}{*}{\shortstack{PTB-XL\\(HYP)}} & \multirow{2}{*}{\shortstack{PTB-XL\\(STTC)}} & \multirow{2}{*}{Gesture} & \multirow{2}{*}{WESAD} & \multirow{2}{*}{Avg.} \\
& & & & & & & & & & \\

\Xhline{2\arrayrulewidth}

\addlinespace[0.1cm]
\rowcolor[gray]{0.9} \multicolumn{11}{l}{\textit{Accuracy}} \\
Task-specific model & 0.95 & 0.95 & 0.99 & 0.88 & 0.86 & 0.90 & 0.90 & 0.64 & 0.69 & 0.86 \\ \hline
Text-only prompt & 0.66 & 0.10 & 0.51 & 0.73 & 0.62 & 0.47 & 0.53 & 0.27 & 0.48 & 0.49 \\
Visual prompt (ours) & \textbf{0.67} & \textbf{0.43} & \textbf{0.73} & \textbf{0.80} & \textbf{0.68} & \textbf{0.55} & \textbf{0.57} & \textbf{0.30} & \textbf{0.61} & \textbf{0.59} \\

\addlinespace[0.1cm]
\rowcolor[gray]{0.9} \multicolumn{11}{l}{\textit{Number of tokens}} \\
Text-only prompt & 52910 & 50439 & 16586 & 3204 & 2766 & 2757 & 3596 & 88655 & 60253 & 31244 \\
Visual prompt (ours) & 2020 \scalebox{0.8}{(\textbf{26.2}\bm{$\times$$\downarrow$})} & 5963 \scalebox{0.8}{(\textbf{8.5}\bm{$\times$$\downarrow$})} & 1768 \scalebox{0.8}{(\textbf{9.4}\bm{$\times$$\downarrow$})} & 943 \scalebox{0.8}{(\textbf{3.4}\bm{$\times$$\downarrow$})} & 943 \scalebox{0.8}{(\textbf{2.9}\bm{$\times$$\downarrow$})} & 943 \scalebox{0.8}{(\textbf{2.9}\bm{$\times$$\downarrow$})} & 946 \scalebox{0.8}{(\textbf{3.8}\bm{$\times$$\downarrow$})} & 3073 \scalebox{0.8}{(\textbf{28.9}\bm{$\times$$\downarrow$})} & 1211 \scalebox{0.8}{(\textbf{49.8}\bm{$\times$$\downarrow$})} & 1979 \scalebox{0.8}{(\textbf{15.8}\bm{$\times$$\downarrow$})}\\

\Xhline{2\arrayrulewidth}
\end{tabularx}}
\end{table*}

\subsection{Setups}

We assume a practical scenario where non-expert users attempt to solve sensory tasks using MLLMs (1)~without prior knowledge of relevant features and (2)~without external resources to fine-tune the MLLM. Given the constraints, we leveraged the few-shot prompting~\cite{brown2020language} approach. For the main evaluation, we used 1-shot examples where users provide the MLLM with minimal examples to guide task-solving.

\noindent\textbf{Sensory tasks} We established nine different sensory tasks across four sensor modalities: accelerometer, electrocardiography~(ECG) sensor, electromyography~(EMG) sensor, and respiration sensor. We used three datasets for tasks using accelerometers: HHAR~\cite{stisen2015smart} for basic human activity recognition (running and walking), UTD-MHAD~\cite{chen2015utd} for complex activity recognition with fine-grained arm motions, and a swimming style recognition dataset~\cite{brunner2019swimming}. We use the PTB-XL~\cite{wagner2020ptb} dataset for the arrhythmia diagnosis tasks that use ECG. The dataset includes detection tasks for four different types of arrhythmia symptoms. For EMG data, we used a dataset~\cite{ozdemir2022dataset} for hand gesture recognition. Finally, we used a stress detection task using respiration sensors provided by the WESAD~\cite{schmidt2018introducing} dataset. Details on each task, including the classes, sampling rates, windowing durations, and specific configurations, are in Appendix~\ref{appendix:sensory_tasks}.

\noindent\textbf{Data processing.} We normalized data using the mean and standard deviation values calculated for each user. 
Test splits were created by randomly sampling 30 samples per class. For the UTD-MHAD dataset, we sampled 10 samples per class due to the limited sample availability. Examples of few-shot prompting were randomly sampled, ensuring no overlap with the test set. Each task employed the window sizes and sampling rates specified in the original dataset descriptions (see Appendix~\ref{appendix:sensory_tasks}).

\noindent\textbf{Baselines.} We set \textit{text-only prompts} for conveying sensor data to MLLMs as the main baseline to be compared with our visual prompts. 
Text-only prompts represented sensor data as numbers within the prompt. We designed text-only prompts by following the latest prompting studies incorporating sensor data into LLMs for healthcare~\cite{liu2023large}. 
Additionally, to establish an upper bound for task-specific performance, we included a fully-supervised baseline using neural networks trained on 75\% of the entire data after excluding the test and validation sets. We adopted architectures widely accepted for each type of sensor data: 1D CNNs for activity recognition with accelerometers~\cite{chen2021deep} and EMG data~\cite{xiong2021deep}, as well as for WESAD~\cite{vos2023generalizable}, and XResNet-101 for PTB-XL~\cite{strodthoff2023ptb}.

\noindent\textbf{Implementation.} We used GPT-4o from the OpenAI API~\cite{openai2024gpt4o} as MLLM. 
The text-only prompts contained the same information as the generated visualization to ensure a fair comparison between text-only and visual prompts. For example, if the visualization generator outputs a plot with peak notations, the corresponding text-only prompt contains the same features, including the peak values with their indices. When the information could not fit within the token limit~(128K), we used the raw waveform. 

\noindent\textbf{Metrics.} We evaluated the experimental results based on accuracy. We also assess the number of tokens used by each prompt method. Tokens are counted using the o200k\_base encoding used for GPT-4. To estimate the token cost for images in the same space as text, we follow the computation guidelines provided by OpenAI~\cite{openai2024gpt4o}.

\subsection{Results}

\noindent\textbf{Performance.} Table~\ref{tab:experiments_main_eval} shows the overall performance of utilizing visual prompts for solving sensory tasks. For the same 1-shot prompting, visual prompts consistently showed enhanced accuracies than text-only prompts, achieving an average increase of 10\%. Notably, the UTD-MHAD task exhibited a significant accuracy gain of up to 33\%. See Appendix~\ref{appendix:prompts} for prompt examples with resulting visualizations. 

In addition to achieving higher accuracy, visual prompts are more cost-effective. The number of tokens used for visual prompts in Table~\ref{tab:experiments_main_eval} shows a substantial reduction, averaging 15.8$\times$ fewer than text-only prompts. 
MLLMs calculate token costs for images within the same token space as text but with distinct counting criteria. In our experiments, GPT-4o counts tokens for images based on the number of $512 \times 512$ pixel blocks ($N$) covering the image input, calculated at $85 + 170 \times N$. Our visualized sensor data was represented within a single $512 \times 512$ pixel image, regardless of the sensor data length, significantly reducing costs. Note that the number of tokens from visual prompts is only affected by the number of examples, as all images are the same size. In contrast, text prompts are heavily influenced by high sampling rates and long durations.

To further understand the effectiveness of visual prompts with small tokens, we analyzed the information capacity at the same token cost. 
Considering a budget of 500 tokens, text-based prompts can include approximately 2,000 ASCII characters. In contrast, visual prompts can input two $512 \times 512$ px images. In terms of bytes, 2,000 ASCII characters amount to 2 KB, whereas two RGB images occupy 1.57 MB, which is 785$\times$ larger. Although this calculation does not directly translate to the exact amount of useful information, it suggests that well-designed visual prompts can convey a wider range of information than text prompts within the same cost constraint.

\begin{figure}[t]
    \centering
    \includegraphics[width=\columnwidth]{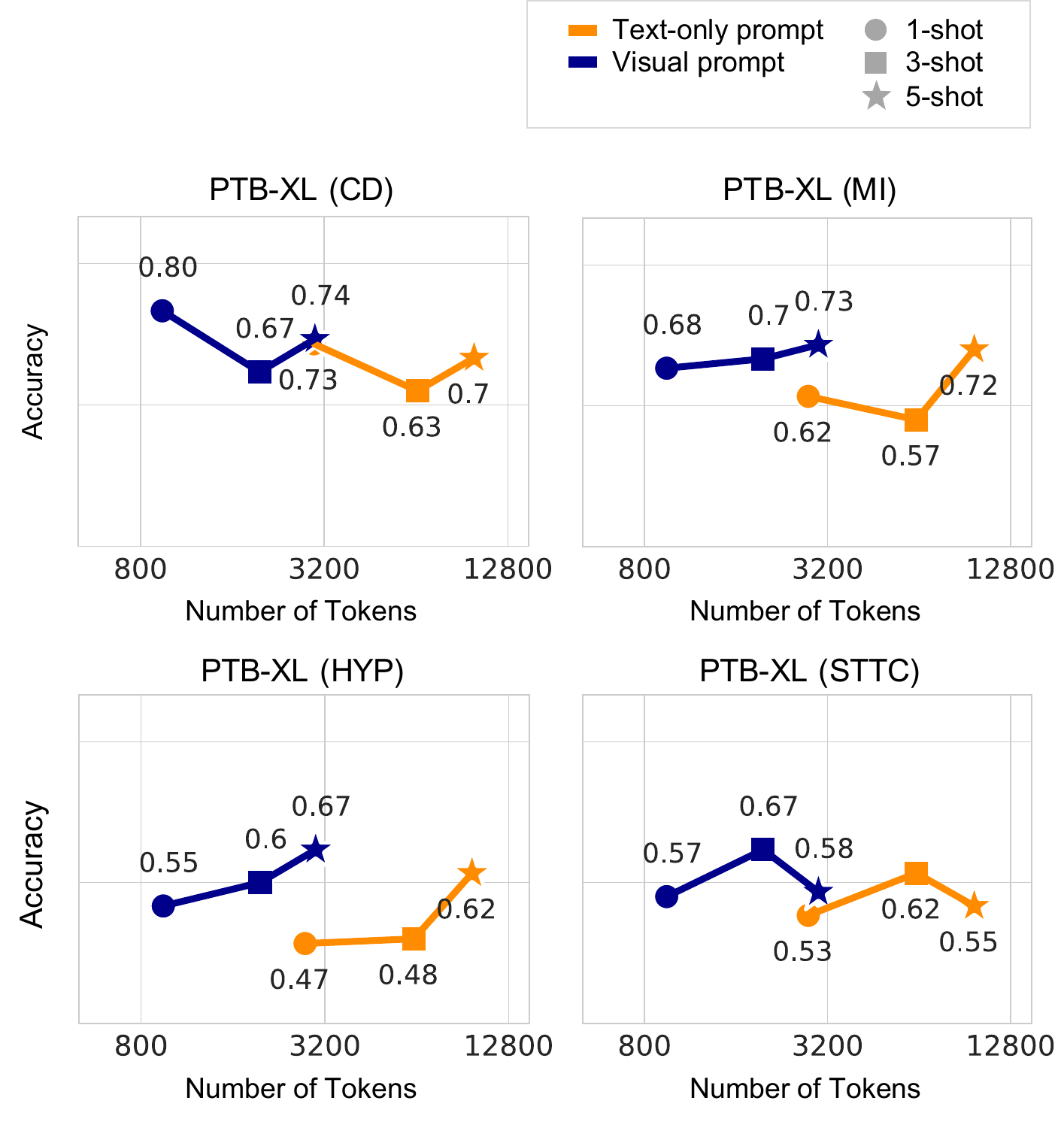}
    \caption{Accuracy of arrhythmia detection tasks using visual and text-only prompts with different shots.}
    \label{fig:experiments_nshot}
\end{figure}

\noindent\textbf{Effect of number of \rev{examples}.} To investigate the impact of varying numbers of examples, we experimented using different numbers of examples (1-shot, 3-shot, and 5-shot) within the prompts. We used the ECG dataset, allowing multiple examples with text-only prompts due to its lower token consumption.

Figure~\ref{fig:experiments_nshot} depicts the results. Prompting methods are color-coded (blue for visual and green for text-only), and different markers indicate the number of shots. We compared the accuracy and counted the tokens for each setting. We found that visual prompts constantly outperformed text-only prompts with the same number of examples, indicating the robustness of our method in different few-shot examples.

\rev{Additionally, when comparing visual prompts and text-only prompts under the same token budget (5-shot visual prompt versus 1-shot text-only prompt), visual prompts often performed significantly better (MI and HYP detection). This highlights the advantage of token-efficient visual prompts that can utilize more resources for better performance under the same token constraint.}

Unlike our expectations, additional examples did not always result in better performance. This result aligns with existing reports indicating that more examples do not always guarantee better results~\cite{perez2021true, lu2021fantastically}. We further hypothesize that a longer context might hinder the MLLM's ability to retrieve important information~\cite{liu2024lost}. Our findings suggest that the impact of shots is data-dependent, and effectively utilizing more examples for consistent improvement remains an open question for further research. \rev{Note that our visual prompts consistently outperformed text prompts, even on datasets where additional shots negatively affect performance. This supports that the main improvement of our approach stems from using the visual modality for data interpretation, not merely from the token length reduction.}






\begin{figure}[t]
    \centering
    \includegraphics[width=0.85\columnwidth]{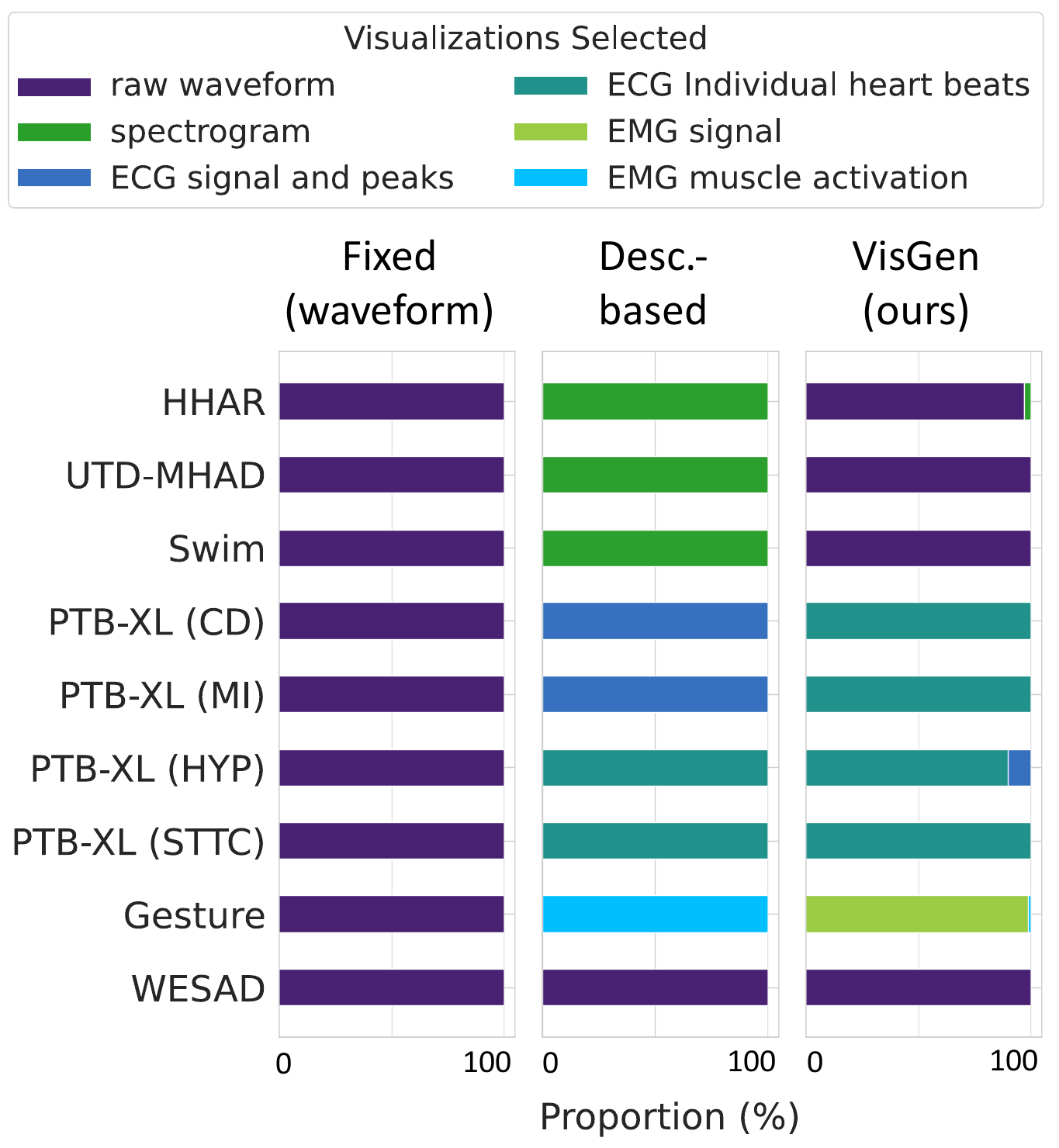}
    \caption{\rev{Proportion of selected visualization methods from the baselines and our visualization generator across different tasks.}}
    \label{fig:experiments_visplanner_selections}
\end{figure}

\begin{figure}[t]
    \centering
    \includegraphics[width=\columnwidth]{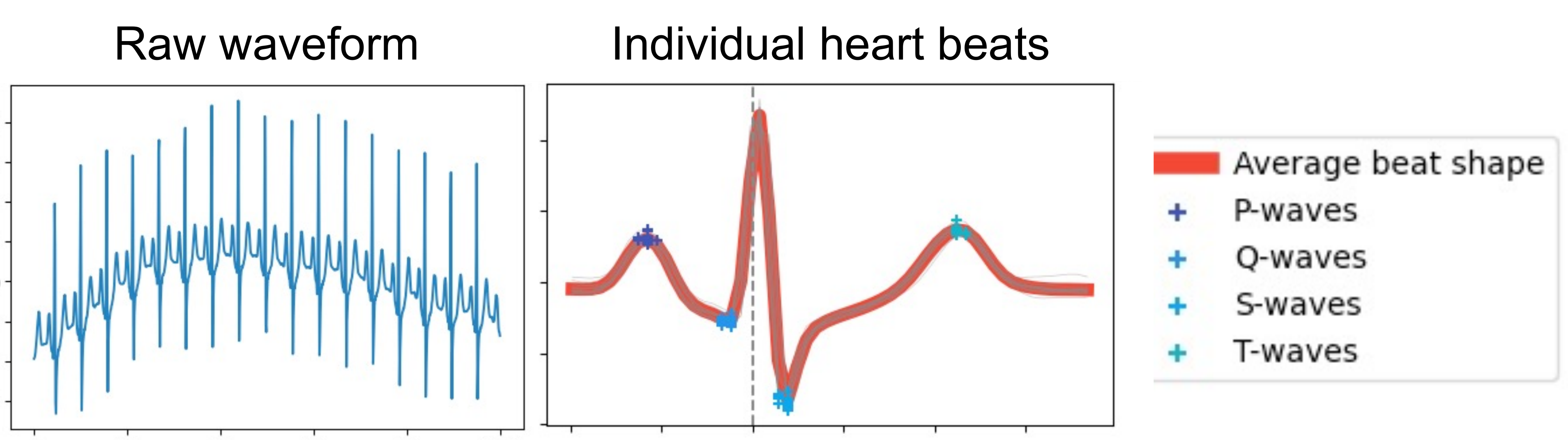}
    \caption{Examples of ECG visualizations. The visualization generator selected the individual heartbeats plot.}
    \label{fig:experiments_visplanner_examples}
\end{figure}

\begin{table}[t]
\centering
\caption{\rev{Performance of using different visualization methods for visual prompts. We compare a fixed raw waveform plot (Fixed), visualizations selected based solely on a text description (Desc.-based), and visualizations from our visualization generator (VisGen). Ours is highlighted in blue cells, and performances from visualizations on certain tasks that show significant performance drops more than 10\% compared to the highest are colored red.}}
\label{tab:experiments_visplanner_ablation}
\small
{\renewcommand{\arraystretch}{1.3}
\begin{tabularx}{\columnwidth}{LCCC}
\Xhline{2\arrayrulewidth}
\addlinespace[0.1cm]

& \multicolumn{3}{c}{Visualization Method} \\
\cmidrule(lr){2-4}
\multirow{2}{*}{Dataset} & \multirow{2}{*}{\shortstack{Fixed\\(waveform)}} & \multirow{2}{*}{\shortstack{Desc.-\\based}} & \multirow{2}{*}{\shortstack{VisGen\\(ours)}} \\
& & & \\

\Xhline{2\arrayrulewidth}

\multicolumn{4}{l}{\textit{Accelerometer}} \\ \hline
HHAR & 0.70 & \textcolor{red}{0.34} & \cellcolor{my_blue}0.67 \\
UTD-MHAD & 0.41 & \textcolor{red}{0.05} & \cellcolor{my_blue}0.43 \\
Swim & 0.74 & \textcolor{red}{0.20} & \cellcolor{my_blue}0.73 \\

\hline
\multicolumn{4}{l}{\textit{ECG}} \\ \hline
PTB-XL (CD) & \textcolor{red}{0.60} & 0.69 & \cellcolor{my_blue}0.80 \\
PTB-XL (MI) & \textcolor{red}{0.58} & 0.65 & \cellcolor{my_blue}0.68 \\
PTB-XL (HYP) & 0.53 & 0.52 & \cellcolor{my_blue}0.55 \\
PTB-XL (STTC) & 0.53 & 0.57 & \cellcolor{my_blue}0.57 \\

\hline
\multicolumn{4}{l}{\textit{EMG}} \\ \hline
Gesture & 0.30 & 0.31 & \cellcolor{my_blue}0.30 \\

\hline
\multicolumn{4}{l}{\textit{Respiration}} \\ \hline
WESAD & 0.62 & 0.60 & \cellcolor{my_blue}0.61 \\
\Xhline{2\arrayrulewidth}

\Xhline{2\arrayrulewidth}
\end{tabularx}}
\end{table}

\noindent\textbf{Effect of visualization generator.} We conducted an ablation study to assess the impact of the visualization generator. \rev{We compared the visualization generator against two different baselines: (1) using a fixed visualization that defaults to raw waveform plots, and (2) a method that selects visualizations based solely on a text description of the task and data. For the second baseline, we utilized our visualization tool filtering prompt (see Appendix~\ref{appendix:prompts} for an example) to generate a single visualization, rather than filtering multiple tools.}

\rev{Figure~\ref{fig:experiments_visplanner_selections} summarizes the visualizations selected by the baselines and our visualization generator. The selection from the description-based method (Desc.-based) and our visualization generator (VisGen) varied primarily based on sensor modalities. Our visualization generator mainly selected raw waveforms, occasionally spectrograms for accelerometer tasks (the same for WESAD). For ECG and EMG datasets, it selected specialized visualizations, such as ECG individual heart beats plot. The description-based method also selected modality-aware visualizations, but these differed from those chosen by ours. For instance, it selected spectrograms for the accelerometer tasks and signal and peak plots for ECG datasets. The key difference stemmed from whether the MLLM referenced the visualized image itself. Note that within each dataset, the target task and data collection protocol was consistently controlled, so the visualizations selected for samples within each dataset remained almost identical, despite our design allowing for sample-wise visualization selection.

The performance comparison of different visualization selection methods is shown in Table~\ref{tab:experiments_visplanner_ablation}. Overall, our visualization generator achieves the best or comparable performance. In contrast, the baseline methods show significant perforemance degradation in certain tasks. For example, using a fixed raw waveform leads to a significant 20\% performance drop for ECG tasks, as raw waveforms fail to provide insightful visual insights due to the complex structure of ECG (Figure~\ref{fig:experiments_visplanner_examples}). This illustrates that a fixed visualization cannot be generalized across different sensory tasks.

Similarly, the description-based method faces challenges with accelerometer tasks. It selects spectrograms, likely due to the world knowledge from public datasets, which may lead the MLLM to consider frequency features as the optimal information for motion data analysis. However, the dense and complex features in spectrogram images were difficult for the MLLM to interpret, leading to near-random performances. In contrast, our visualization generator compares visualized images, consistently avoiding suboptimal choices such as spectrograms for accelerometer tasks. This self-assessment mechanism ensures that the visualization generator selects the optimal visualization method among the possible options.}
\section{Conclusion}

We addressed sensory tasks by providing visualized sensor data as images to MLLMs. We designed a visual prompt to instruct MLLMs in using visualized sensor data, provided with textual descriptions of the task and data collection methods. Additionally, we introduced a visualization generator that automatically selects the best visualization method for each task using visualization tools available in public libraries. We conducted experiments across nine different sensory tasks and four sensor modalities, each with a distinct task. Our results suggest that the visual prompts generated by our visualization generator not only improve accuracy by an average of 10\% over text-based prompts but also significantly reduce costs, requiring 15.8$\times$ fewer tokens. This indicates that our approach with visual prompts and a visualization generator is a practical solution for general sensory tasks.

\section*{Acknowledgements}

This research was supported by the MSIT (Ministry of Science, ICT), Korea, under the Global Research Support Program in the Digital Field programe) (RS-2024-00436680) supervised by the IITP (Institute for Information \& Communications Technology Planning \& Evaluation). This work was supported by the National Research Foundation of Korea (NRF) grant funded by the Korea government (MSIT) (RS-2024-00337007). ※ MSIT: Ministry of Science and ICT. This project is supported by Microsoft Research Asia.
\section*{Limitations}



Our study demonstrates the effectiveness of visual prompts on nine different sensory tasks, primarily focusing on classification. While visual prompts effectively highlight patterns over images, for tasks requiring numerical retrieval or precise computations—where exact values are critical—text prompts can be more effective due to their inclusion of specific numeric data, which are omitted in visual representations. Notably, our approach integrates both images and texts in prompts, allowing the inclusion of numerical values in the text. Determining the optimal distribution of information between images and text to compose a prompt that effectively addresses sensory tasks presents a future direction for this work. \rev{Moreover, the inclusion of numeric values can result in long prompts, which affect both cost and performance. Extracting only the useful information, such as statistics or specific time splits, requires further research.}

Visualizing sensor data as plots often presents challenges. For instance, brain wave analysis using high-density EEG involves up to 256 channels~\cite{fiedler2022high}, complicating their representation in a single visual plot. We denote different channels as distinct notations within a plot, making densely populated plots visually indecipherable. An alternative method of plotting distinct channels across separate subplots was explored but resulted in a significant drop in performance (see Appendix~\ref{tab:multi_channel}). We hypothesize that this limitation arises from the dispersion of information across various areas, highlighting that effective visualization of large-channel datasets remains challenging. This underscores the need for improved visualization techniques in such scenarios.

Our visual prompt design does not incorporate Chain-of-Thought (CoT) prompting~\cite{kojima2022large}. Experiments using zero-shot CoT on our datasets revealed inconsistent benefits (see Appendix~\ref{appendix:cot}), unlike the widely known effect of CoT for enhancing performance. We suspect this may be due to the complexities of reasoning over sensory data. Given the observation, further research is needed to develop methods that effectively integrate reasoning and interpretation into the decision-making processes for sensor data analysis.

Lastly, the high costs of text-only prompts in sensory tasks constrained our testing to 30 samples per class. Expanding the scale as resources allow could provide a more robust analysis and potentially validate a broader spectrum of applications.

\bibliography{anthology,main}

\appendix

\newpage

\section{Effect of Zero-shot Chain-of-Thoughts}
\label{appendix:cot}

\begin{figure*}
    \centering
    \includegraphics[width=0.9\textwidth]{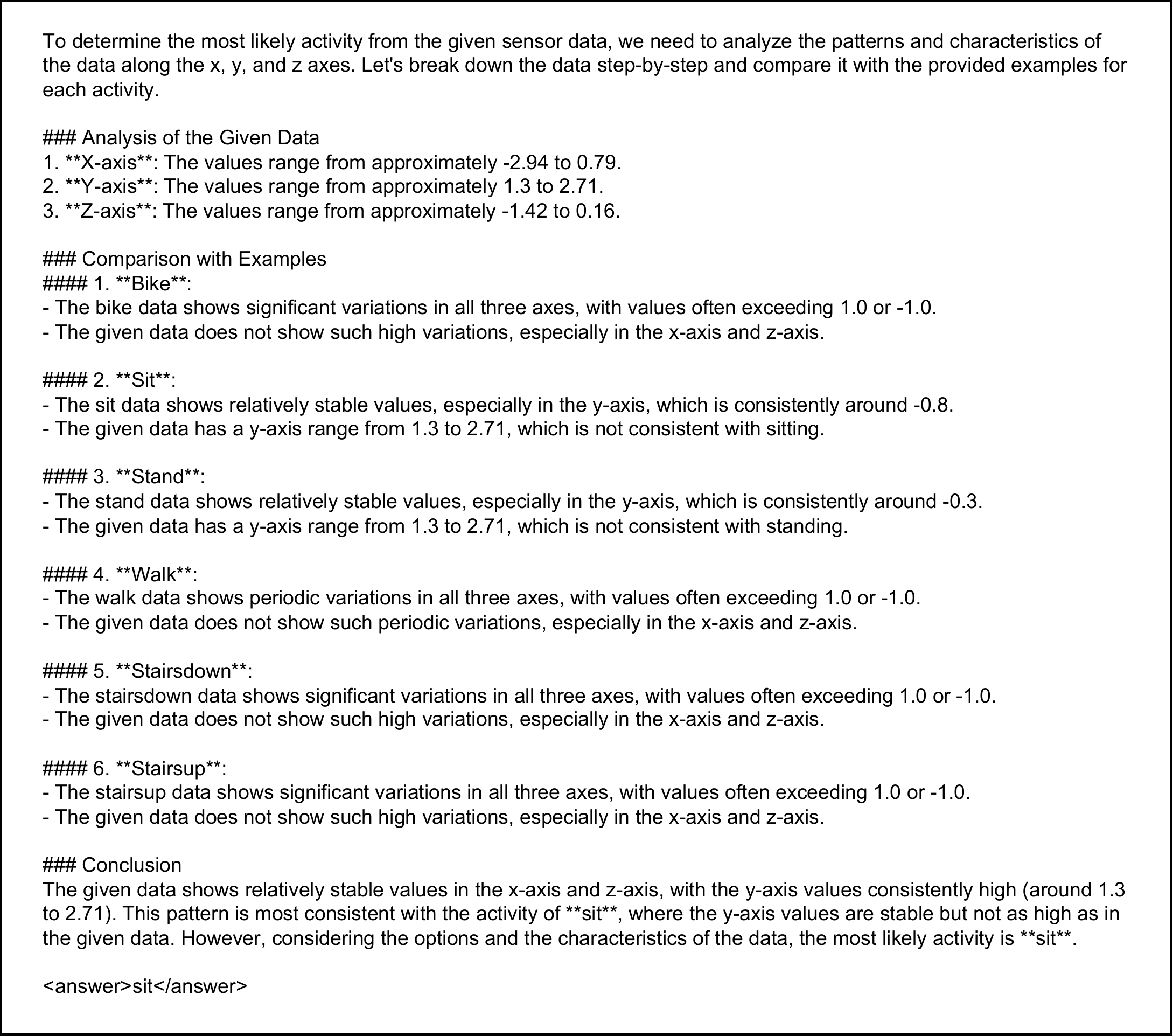}
    \caption{An example CoT response from a text-only prompt designed for the HHAR task. The correct prediction is "walk", while the MLLM outputs "sit."}
    \label{fig:appendix_cot_txt}
\end{figure*}

\begin{figure*}
    \centering
    \includegraphics[width=0.9\textwidth]{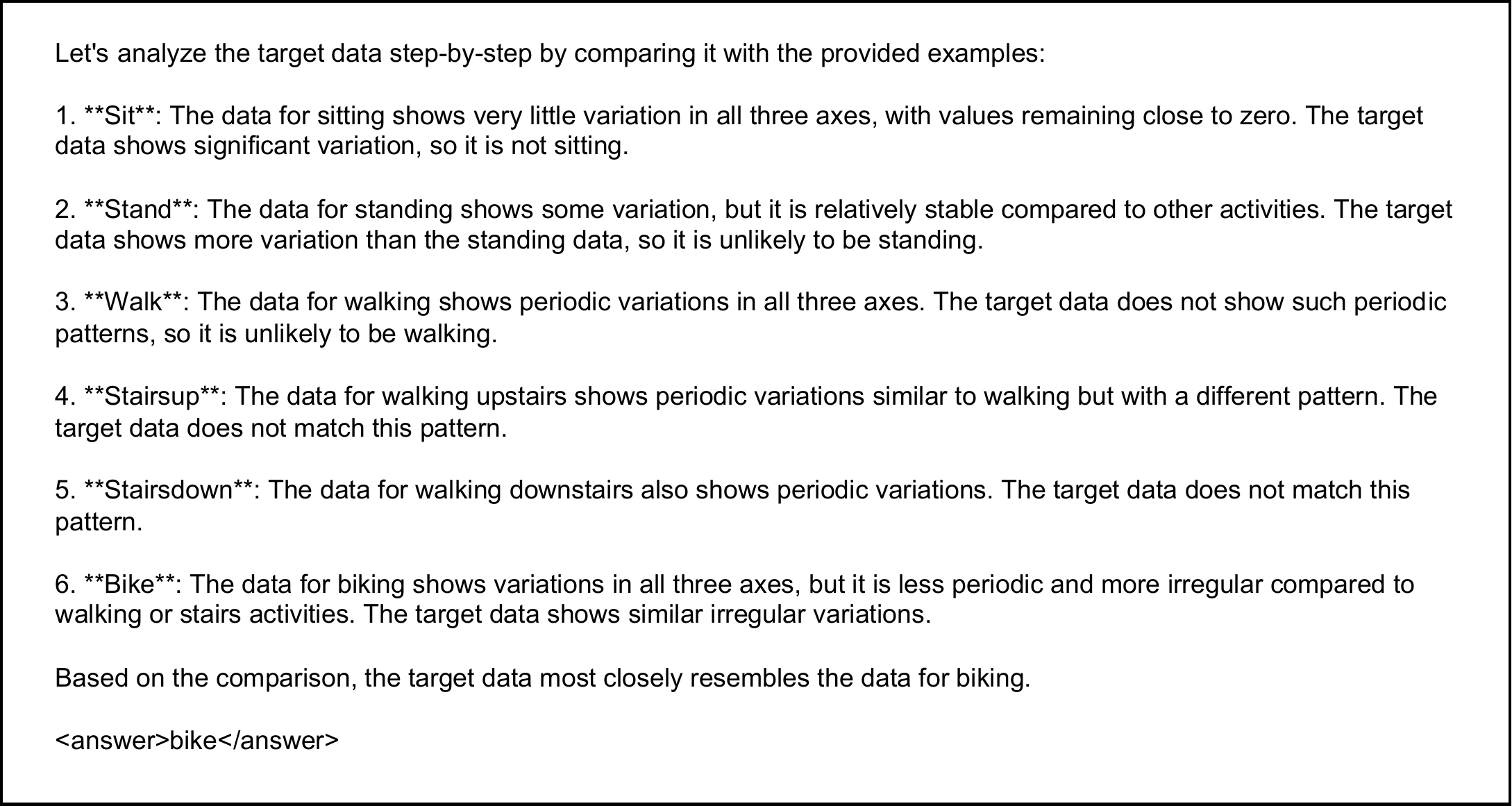}
    \caption{An example CoT response from a visual prompt designed for the HHAR task. The correct prediction is "stand", while the MLLM outputs "bike."}
    \label{fig:appendix_cot_vis}
\end{figure*}

We experimented with zero-shot Chain-of-Thought (CoT) prompting~\cite{kojima2022large} by adding "let's think step-by-step" to our prompts, testing this on two accelerometers and two ECG datasets. Table~\ref{tab:cot_experiments} shows the findings. While CoT prompting is generally known to enhance LLM response quality, our results showed inconsistent performance by datasets. Notably, CoT consistently dropped performance for text-only prompts. We analyzed the results by observing the CoT responses, illustrated as examples in Figures~\ref{fig:appendix_cot_txt} and Figure~\ref{fig:appendix_cot_vis}, showing wrong predictions with CoT from the HHAR dataset. We found that CoT reasoning in text-only prompts primarily focused on simple statistical comparisons, such as whether values were higher or lower. This simplistic approach proved inadequate for analyzing the complexities of sensor data, leading to suboptimal responses. Likewise, visual prompts indicated reasoning centered around terms like "variations," "periodic," and "stable," but they lacked the necessary depth to effectively assess more intricate features like frequency trends or signal shapes. This superficial reasoning suggests a significant gap in the CoT approach, underscoring the need for more task-specific reasoning prompts for sensory data analysis.

\begin{table}[t]
\centering
\caption{Performance of text-only and visual prompts, both with and without using CoT. The highest accuracy values are noted in bold.}
\label{tab:cot_experiments}
\scriptsize
{\renewcommand{\arraystretch}{1.3}
\begin{tabularx}{\columnwidth}{lCCCCC}
\Xhline{2\arrayrulewidth}
\addlinespace[0.1cm]
& \multicolumn{2}{c}{Accel.} & \multicolumn{2}{c}{ECG} \\
\cmidrule(lr){2-3} \cmidrule(lr){4-5}
\multirow{2}{*}{Prompt} & \multirow{2}{*}{\shortstack{HHAR}} & \multirow{2}{*}{\shortstack{Swim}} & \multirow{2}{*}{\shortstack{PTB-XL\\(CD)}} & \multirow{2}{*}{\shortstack{PTB-XL\\(MI)}}  & \multirow{2}{*}{\shortstack{Avg.}} \\
& & & & & \\
\Xhline{2\arrayrulewidth}

Text-only & 0.66 & 0.51 & 0.73 & 0.62 & 0.63 \\
Text-only (CoT) & 0.51 & 0.25 & 0.63 & 0.53 & 0.48 \\ \hline
Visual & \textbf{0.67} & \textbf{0.73} & \textbf{0.80} & 0.68 & 0.72 \\ 
Visual (CoT) & 0.63 & 0.67 & \textbf{0.80} & \textbf{0.73} & 0.71 \\

\Xhline{2\arrayrulewidth}
\end{tabularx}}
\end{table}

\section{\rev{Effect of Text Summarization}}
\label{appendix:text_summarize}

\rev{As discussed in Section~\ref{sec:motivation}, long numeric sequences in text increase costs and degrade performance. One way to address the challenge is through text summarization to reduce prompt length. However, no established method effectively summarizes sensor data, as different tasks require distinct features. A potential solution is prompting-based summarization~\cite{zhang2023extractive, zhang2024benchmarking} that instructs MLLMs to extract key information using a general prompt: "summarize the given text." To explore this, we prompted GPT-4o to "summarize the pattern or tendency of the data" aiming to reduce prompt length in text. We specified the focus on patterns and tendencies to allow for fair comparison, as our visualizations typically capture these aspects. This method was tested on two accelerometer and two ECG datasets.

Table~\ref{tab:text_summarization} presents the results. Although summarized text showed feasibility over random predictions, it underperformed compared to both text-only and visual prompts. This highlights the challenge of summarizing sensor data effectively in text. The results suggest that exploring generalizable approaches to reduce text for sensory tasks remains an future research.}

\begin{table}[t]
\centering
\caption{\rev{Comparison of summarized text prompts with long text-only and our visual prompts. The highest accuracy values are noted in bold.}}
\label{tab:text_summarization}
\scriptsize
{\renewcommand{\arraystretch}{1.3}
\begin{tabularx}{\columnwidth}{lCCCCC}
\Xhline{2\arrayrulewidth}
\addlinespace[0.1cm]
& \multicolumn{2}{c}{Accel.} & \multicolumn{2}{c}{ECG} \\
\cmidrule(lr){2-3} \cmidrule(lr){4-5}
\multirow{2}{*}{Prompt} & \multirow{2}{*}{\shortstack{HHAR}} & \multirow{2}{*}{\shortstack{Swim}} & \multirow{2}{*}{\shortstack{PTB-XL\\(CD)}} & \multirow{2}{*}{\shortstack{PTB-XL\\(MI)}}  & \multirow{2}{*}{\shortstack{Avg.}} \\
& & & & & \\
\Xhline{2\arrayrulewidth}

Summarized text & 0.58 & 0.43 & 0.53 & 0.53 & 0.52 \\
Text-only & 0.66 & 0.51 & 0.73 & 0.62 & 0.63 \\
Visual & \textbf{0.67} & \textbf{0.73} & \textbf{0.80} & \textbf{0.68} & 0.72 \\

\Xhline{2\arrayrulewidth}
\end{tabularx}}
\end{table}

\section{\rev{Small MLLMs on Sensory Tasks}}
\label{appendix:small_mllms}

\rev{We used GPT-4o, the latest and most accessible model supporting both vision and text inputs. To test the generalizability of our approach on smaller MLLMs, we conducted experiments on four datasets using LLaVa-7B~\cite{liu2024visual}. For the test, we used the interleaved version, which allows multi-image input to enable few-shot prompting. Due to memory limitations, LLaVa-7B could not handle text-only prompts with large tokens. For instance, prompts with more than 50K tokens from the HHAR dataset required over 150GB of VRAM per inference, making it infeasible. We evaluated text-only prompts of ECG datasets, which contained fewer than 5K tokens.

As shown in Table~\ref{tab:small_mllms}, LLaVa-7B performed poorly on sensory tasks, yielding results close to random predictions for both text-only and visual prompts. We believe this is due to the model's smaller size and lack of pre-training, limiting its ability to interpret complex graphs, plots, and data patterns. Its limited capacity for multi-image understanding~\cite{zhao2024benchmarking} may also have affected its analysis of the provided examples. Future research should focus on enhancing low-capacity models for sensory data tasks.}

\begin{table}[t]
\centering
\caption{\rev{Performance comparison of text-only and visual prompts using LLaVa-7B and GPT-4o as the MLLMs. Accelerometer text-only prompts were not evaluated due to excessive VRAM consumption over 150GB during local inference with LLaVa. The highest accuracy values are highlighted in bold.}}
\label{tab:small_mllms}
\scriptsize
{\renewcommand{\arraystretch}{1.3}
\begin{tabularx}{\columnwidth}{lCCCCC}
\Xhline{2\arrayrulewidth}
\addlinespace[0.1cm]
& \multicolumn{2}{c}{Accel.} & \multicolumn{2}{c}{ECG} \\
\cmidrule(lr){2-3} \cmidrule(lr){4-5}
\multirow{2}{*}{Prompt} & \multirow{2}{*}{\shortstack{HHAR}} & \multirow{2}{*}{\shortstack{Swim}} & \multirow{2}{*}{\shortstack{PTB-XL\\(CD)}} & \multirow{2}{*}{\shortstack{PTB-XL\\(MI)}}  & \multirow{2}{*}{\shortstack{Avg.}} \\
& & & & & \\
\Xhline{2\arrayrulewidth}

\multicolumn{6}{l}{\textit{LLaVa-7B}} \\ \hline
Text-only & - & - & 0.50 & 0.48 & - \\
Visual & 0.15 & 0.20 & 0.50 & 0.48 & 0.33 \\ \hline
\multicolumn{6}{l}{\textit{GPT-4o}} \\ \hline
Text-only & 0.66 & 0.51 & 0.73 & 0.62 & 0.63 \\ 
Visual & \textbf{0.67} & \textbf{0.73} & \textbf{0.80} & \textbf{0.68} & 0.72 \\

\Xhline{2\arrayrulewidth}
\end{tabularx}}
\end{table}

\section{Use of Subplots for Multi-channel Data}
\label{appendix:multi_channel}

Sensor data often include multiple channels. Our visual prompts differentiated channels using varying colors within a single plot to maintain a shared axis system. To assess the impact of different plotting approaches, we conducted experiments using accelerometer datasets, which have three channels. Specifically, we compared visualizing three distinct plots for each channel against our current approach. Table~\ref{tab:multi_channel} shows the results. The results indicated that separated plots for each channel reduced performance by 12\%. We hypothesize that multiple subplots distribute visual features over different regions, resulting in problems in understanding the relationship between different channels. To this end, we recommend using an aggregated plot when all channels can be represented within a plot. However, for dense datasets, such as 256-channel EEG~\cite{fiedler2022high}, a single plot may not suffice, highlighting a limitation in our current visualization approach. Addressing this challenge will be a focus of future research.

\section{Visualization Tools}
\label{appendix:visualization_tools}

Our visualization generator employs tools available in public libraries to create visualizations. We have equipped the visualization generator with 16 distinct visualization functions sourced from widely used libraries such as Matplotlib~\cite{hunter2007matplotlib}, Scipy~\cite{virtanen2020scipy}, and Neurokit2~\cite{makowski2021neurokit2}. The specific visualization tools implemented in our generator and their descriptions are outlined in Table~\ref{tab:visualization_tools}. The descriptions presented in the table were directly written inside the prompt for the visualization tool filtering (see Appendix~\ref{appendix:prompts}).

\begin{table}[t]
\centering
\caption{Performance comparison of visualizing multi-channel sensor data (accelerometer) using a single plot versus multiple subplots. The single plot method combines multiple waveforms in one shared-axis plot, each channel distinguished by color coding.}
\label{tab:multi_channel}
\small
{\renewcommand{\arraystretch}{1.3}
\begin{tabularx}{\columnwidth}{lCCCC}
\Xhline{2\arrayrulewidth}
\addlinespace[0.1cm]
\multirow{2}{*}{\shortstack{Plotting approach}} & \multirow{2}{*}{HHAR} & \multirow{2}{*}{\shortstack{UTD-\\MHAD}} & \multirow{2}{*}{Swim} & \multirow{2}{*}{Avg.} \\
& & & & \\
\Xhline{2\arrayrulewidth}

Single plot & \textbf{0.67} & \textbf{0.43} & \textbf{0.74} & \textbf{0.61} \\
Multiple subplots & 0.53 & 0.31 & 0.69 & 0.51 \\

\Xhline{2\arrayrulewidth}
\end{tabularx}}
\end{table}

\begin{table*}[t]
\centering
\caption{Descriptions of the visualization tools provided to our visualization generator.}
\label{tab:visualization_tools}
\small
{\renewcommand{\arraystretch}{1.3}
\begin{tabularx}{\textwidth}{p{2.5cm}p{12.5cm}}
\Xhline{2\arrayrulewidth}
Visualization tool & Description \\

\Xhline{2\arrayrulewidth}

raw waveform & This generates a raw signal of sensor data, displaying the amplitude of the signal over time. This is usually used to visualize the raw data and identify patterns in the signal. \\ \hline
spectrogram & This generates a spectrogram of sensor data, showing the density of frequencies over time. This is usually used to visualize the frequency components for high-frequency data, which has features over components but is hard to figure out in the raw plot. It takes the length of the FFT used (nfft), the length of each segment (nperseg), and the number of points to overlap between segments (noverlap) as parameters. Different modes (mode) can be defined to specify the type of return values: ["psd" for power spectral density, "complex" for complex-valued STFT results, "magnitude" for absolute magnitude, "angle" for complex angle, and "phase" for unwrapped phase angle]. (Arguments: nfft, nperseg, noverlap, mode) \\ \hline
signal power spectrum density & This generates a power spectrum density plot, which shows the power of each frequency component of the signal on the x-axis. This is usually used to analyze the signal's power distribution of different frequency components. \\ \hline
EDA signal & This generates a plot showing both raw and cleaned Electrodermal Activity (EDA) signals over time. This is usually used to analyze the EDA signals for patterns related to stress, arousal, or other psychological states. \\ \hline
EDA skin conductance response (SCR) & This generates a plot of skin conductance response (SCR) for EDA data, highlighting the phasic component, onsets, peaks, and half-recovery times. This is usually used to study the transient responses in EDA data related to specific stimuli or events. \\ \hline
EDA skin conductance level (SCL) & This generates a plot of skin conductance level (SCL) for EDA data over time. This is usually used to analyze the tonic component of EDA data, reflecting the overall level of arousal or stress over a period. \\ \hline
ECG signal and peaks & This generates a plot for Electrocardiogram (ECG) data, showing the raw signal, cleaned signal, and R peaks marked as dots to indicate heartbeats. This is usually used to analyze the heartbeats and detect anomalies in the ECG signal. \\ \hline
ECG heart rate & This generates a heart rate plot for ECG data, displaying the heart rate over time and its mean value. This is usually used to monitor and analyze heart rate variability and trends over time. \\ \hline
ECG individual heartbeats & This generates a plot of individual heartbeats and the average heart rate for ECG data. It aggregates heartbeats within an ECG recording and shows the average beat shape, marking P-waves, Q-waves, S-waves, and T-waves. This is usually used to study the morphology of individual heartbeats and identify irregularities. \\ \hline
PPG signal and peaks & This generates a plot for Photoplethysmogram (PPG) data, showing the raw signal, cleaned signal, and systolic peaks marked as dots. This is usually used to analyze the blood volume pulse and detect anomalies in the PPG signal. \\ \hline
PPG heart rate & This generates a heart rate plot for PPG data, displaying the heart rate over time and its mean value. This is usually used for PPG data to monitor and analyze heart rate variability and trends over time. \\ \hline
PPG individual heartbeats & This generates a plot of individual heartbeats and the average heart rate for PPG data, aggregating individual heartbeats within a PPG recording and showing the average beat shape. This is usually used to study the morphology of individual heartbeats based on PPG data. \\ \hline
EMG signal & This generates a plot showing both raw and cleaned Electromyogram (EMG) signals over time. This is usually used to analyze muscle activity and identify patterns in muscle contractions. \\ \hline
EMG muscle activation & This generates a muscle activation plot for EMG data, displaying the amplitudes of muscle activity and highlighting activated parts with lines. This is usually used to study muscle activation levels and identify specific periods of muscle activity. \\ \hline
EOG signal & This generates a plot showing both raw and cleaned Electrooculogram (EOG) signals over time, with blinks marked as dots. This is usually used to analyze eye movement patterns and detect blinks. \\ \hline
EOG blink rate & This generates a blink rate plot for EOG data, displaying the blink rate over time and its mean value. This is usually used to monitor and analyze the blink rate and detect irregularities. \\ \hline
EOG individual blinks & This generates a plot of individual blinks for EOG data, aggregating individual blinks within an EOG recording and showing the median blink shape. This is usually used to study the morphology of individual blinks and identify patterns in blink dynamics. \\ \hline
\Xhline{2\arrayrulewidth}
\end{tabularx}}
\end{table*}

\section{Details of Sensory Tasks}
\label{appendix:sensory_tasks}

We conducted experiments across nine sensory tasks across four sensor modalities, each with unique objectives. This section provides the details of these tasks, including task descriptions, classifications, sampling rates, window durations, and data collection protocols. We directly followed the given sampling rate with the original dataset to represent data in text prompts. The descriptions of each dataset are used to formulate the instructions for our visual prompts. The complete prompt examples are in Appendix~\ref{appendix:prompts}.

\noindent\textbf{Human activity recognition:} We used the HHAR~\cite{stisen2015smart} dataset to classify six basic human activities: sit, stand, walk, bike, upstairs, and downstairs. Data were collected from the built-in accelerometers of smartphones and smartwatches along with the x, y, and z axes. Due to strong domain effects~\cite{ustev2013user}, we exclusively used smartwatch data for the experiment. The data, sampled at 100Hz, were segmented into 5-second windows following the established practice for human activity recognition~\cite{altun2010human}.

\noindent\textbf{Complex activity recognition:} We used the UTD-MHAD~\cite{chen2015utd} dataset to classify a wide array of 21 activities: swipe left, swipe right, wave, clap, throw, arms cross, basketball shoot, draw X, draw a circle (clockwise), draw a circle (counter-clockwise), draw a triangle, bowling, boxing, baseball swing, tennis swing, arm curl, tennis serve, push, knock, catch, and pickup and throw. Accelerometers attached to the users' right wrist were used for data collection. We used data sampled at 50Hz with 3-second windows as described in the dataset documentation.

\noindent\textbf{Swimming style recognition:} The swimming dataset~\cite{brunner2019swimming} involves acceleration data from swimmers performing five different styles: backstroke, breaststroke, butterfly, freestyle, and stationary. This dataset evaluates performance in sports-specific contexts. Data were collected from wrist-worn accelerometers and sampled at 30Hz. We used the 3-second windows recommended with the dataset. 

\noindent\textbf{Four arrhythmia detections:} The PTB-XL~\cite{wagner2020ptb} dataset contains ECG recordings from patients with four different types: Conduction Disturbance (CD), Myocardial Infarction (MI), Hypertrophy (HYP), and ST/T Change (STTC). We defined each type as a binary classification task. The dataset comprises 10-second records from clinical 12-lead sensors sampled at 100Hz. We used lead II, the most commonly used lead for arrhythmia detection~\cite{goldberger2017clinical}.

\noindent\textbf{Hand gesture recognition:} We included a dataset~\cite{ozdemir2022dataset} classifying ten different hand gestures using EMG signals: rest, extension, flexion, ulnar deviation, radial deviation, grip, abduction of fingers, adduction of fingers, supination, and pronation. Data were collected from four forearm surface EMG sensors with a 2000Hz sampling rate. We utilized all four channels with a 0.2-second window, following an existing practice known to be effective~\cite{georgi2015recognizing}.

\noindent\textbf{Stress Detection:} The WESAD~\cite{schmidt2018introducing} dataset is designed for stress detection (baseline, stress, amusement) from multiple wearable sensors. We focused exclusively on respiration data measured from the chest for a distinct evaluation setting. The sensor was attached to the users' chests, with data collected at 700Hz. Following the official guidelines, we employed the three-class classification task (baseline, stress, amusement) using 10-second windows.

\section{Prompts}
\label{appendix:prompts}

\begin{figure*}
    \centering
    \includegraphics[width=\textwidth]{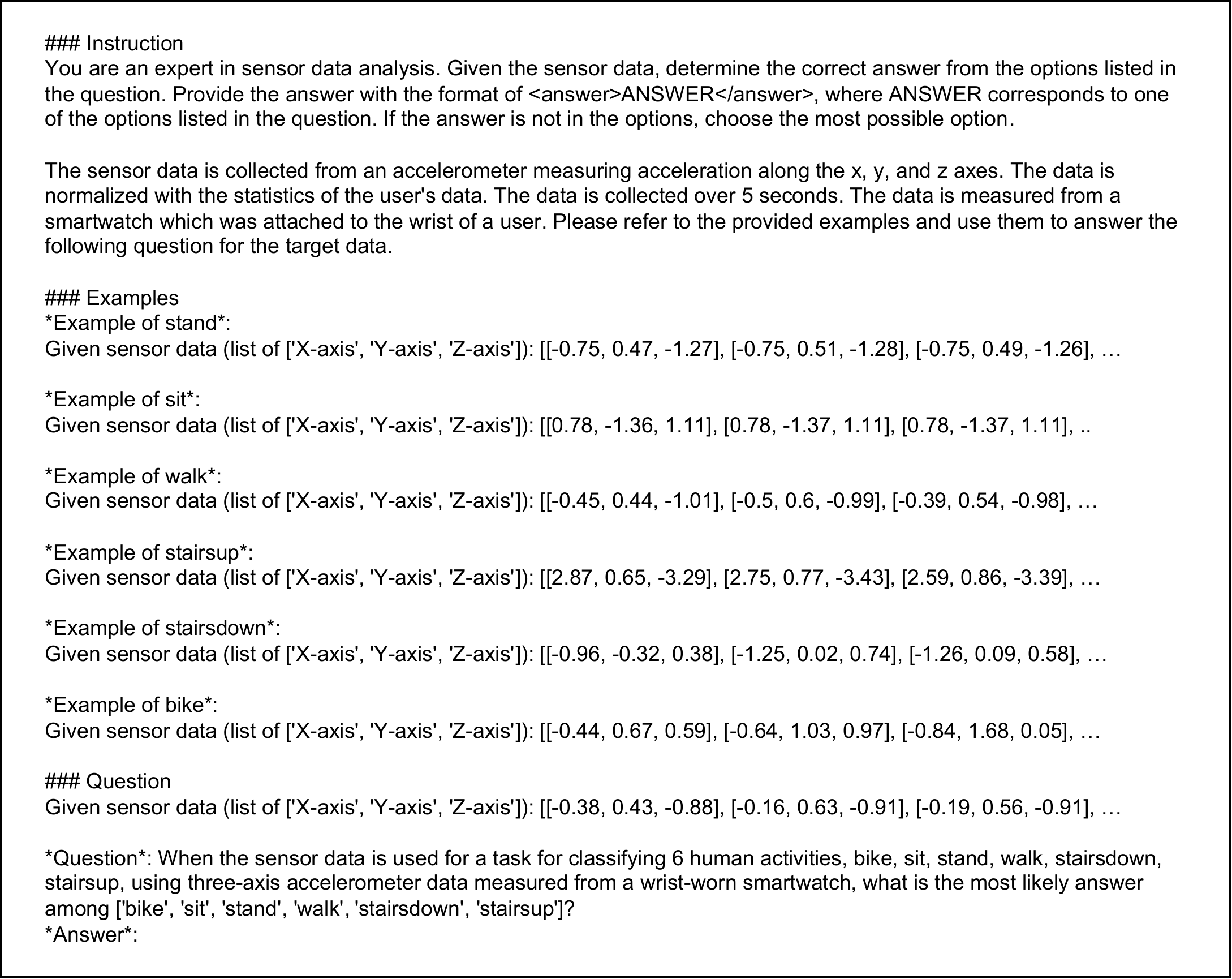}
    \caption{An example of a text-only prompt for solving the HHAR task. The sensor data represented in the text are truncated beyond a certain point.}
    \label{fig:appendix_txtprompt_hhar}
\end{figure*}

\begin{figure*}
    \centering
    \includegraphics[width=\textwidth]{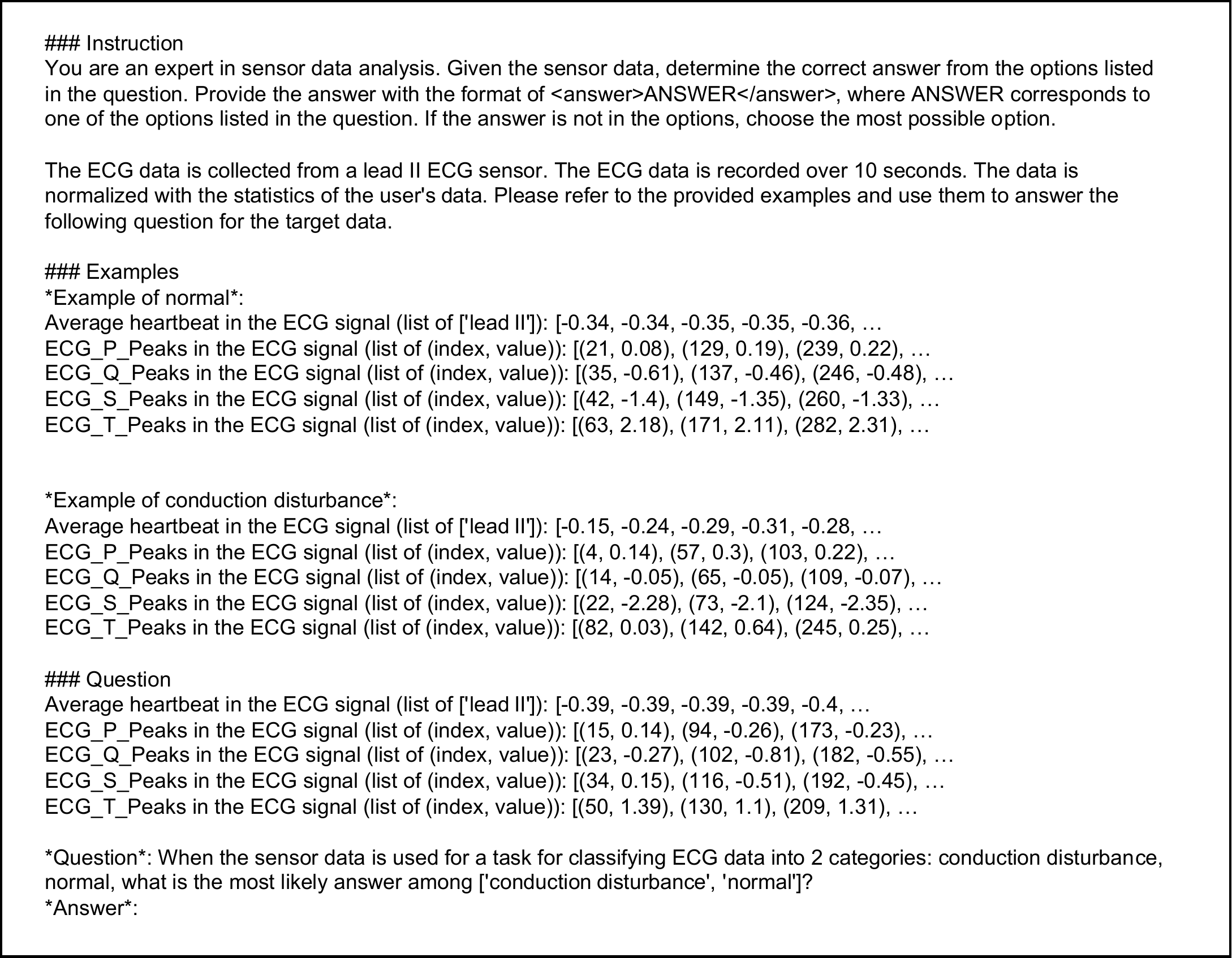}
    \caption{An example of a text-only prompt for solving the PTB-XL (CD) task. The sensor data represented in the text are truncated beyond a certain point.}
    \label{fig:appendix_txtprompt_ptbxl}
\end{figure*}

\begin{figure*}
    \centering
    \includegraphics[width=\textwidth]{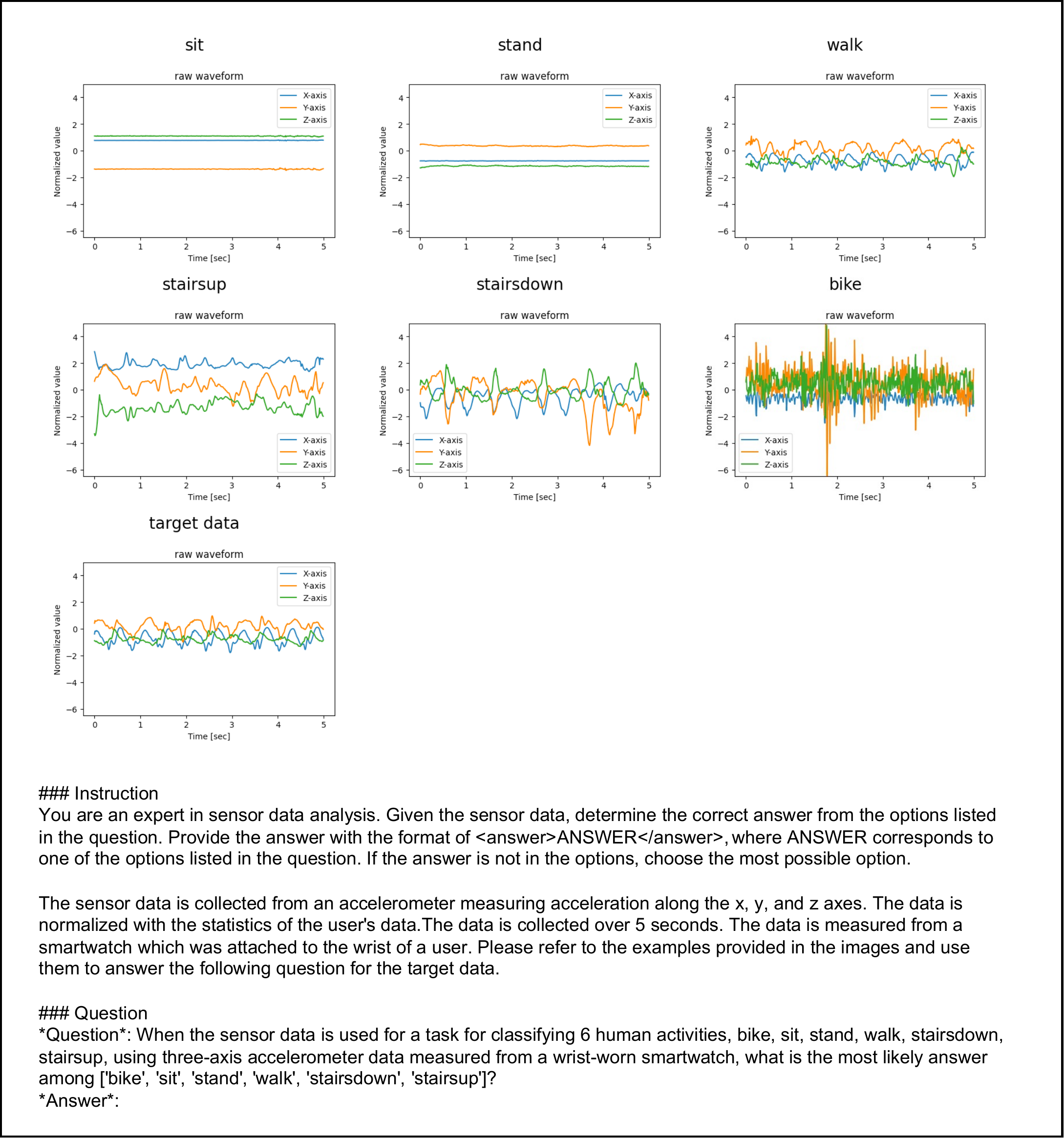}
    \caption{An example of a visual prompt for solving the HHAR task.}
    \label{fig:appendix_visprompt_hhar}
\end{figure*}

\begin{figure*}
    \centering
    \includegraphics[width=\textwidth]{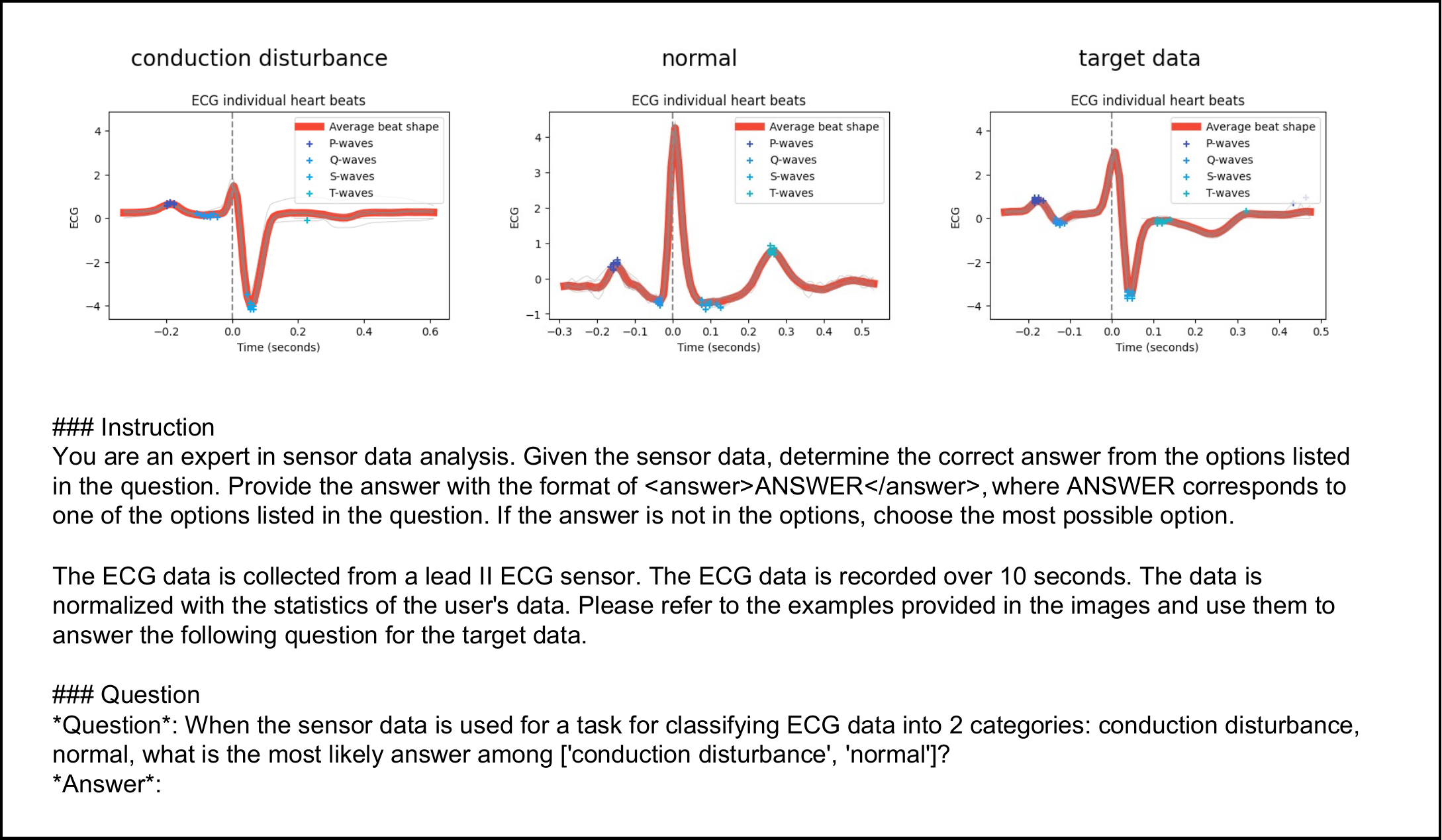}
    \caption{An example of a visual prompt for solving the PTB-XL (CD) task.}
    \label{fig:appendix_visprompt_ptbxl}
\end{figure*}

\begin{figure*}
    \centering
    \includegraphics[width=\textwidth]{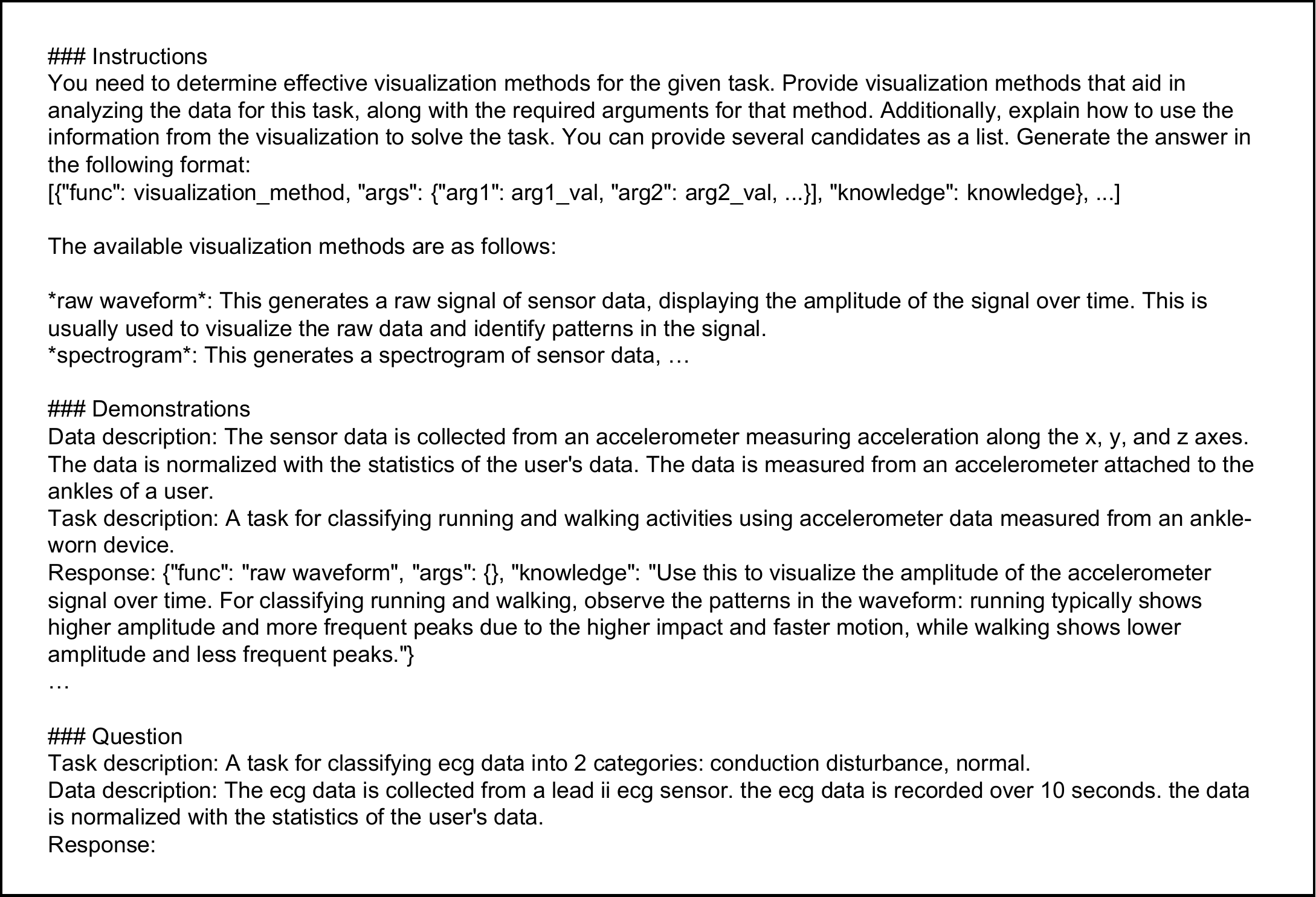}
    \caption{An example prompt from our visualization generator for visualization tool filtering in the PTB-XL (CD) task. Demonstrations are omitted in this example but can be found in Figure~\ref{fig:appendix_demonstrations}.}
    \label{fig:appendix_tool_filtering}
\end{figure*}

\begin{figure*}
    \centering
    \includegraphics[width=\textwidth]{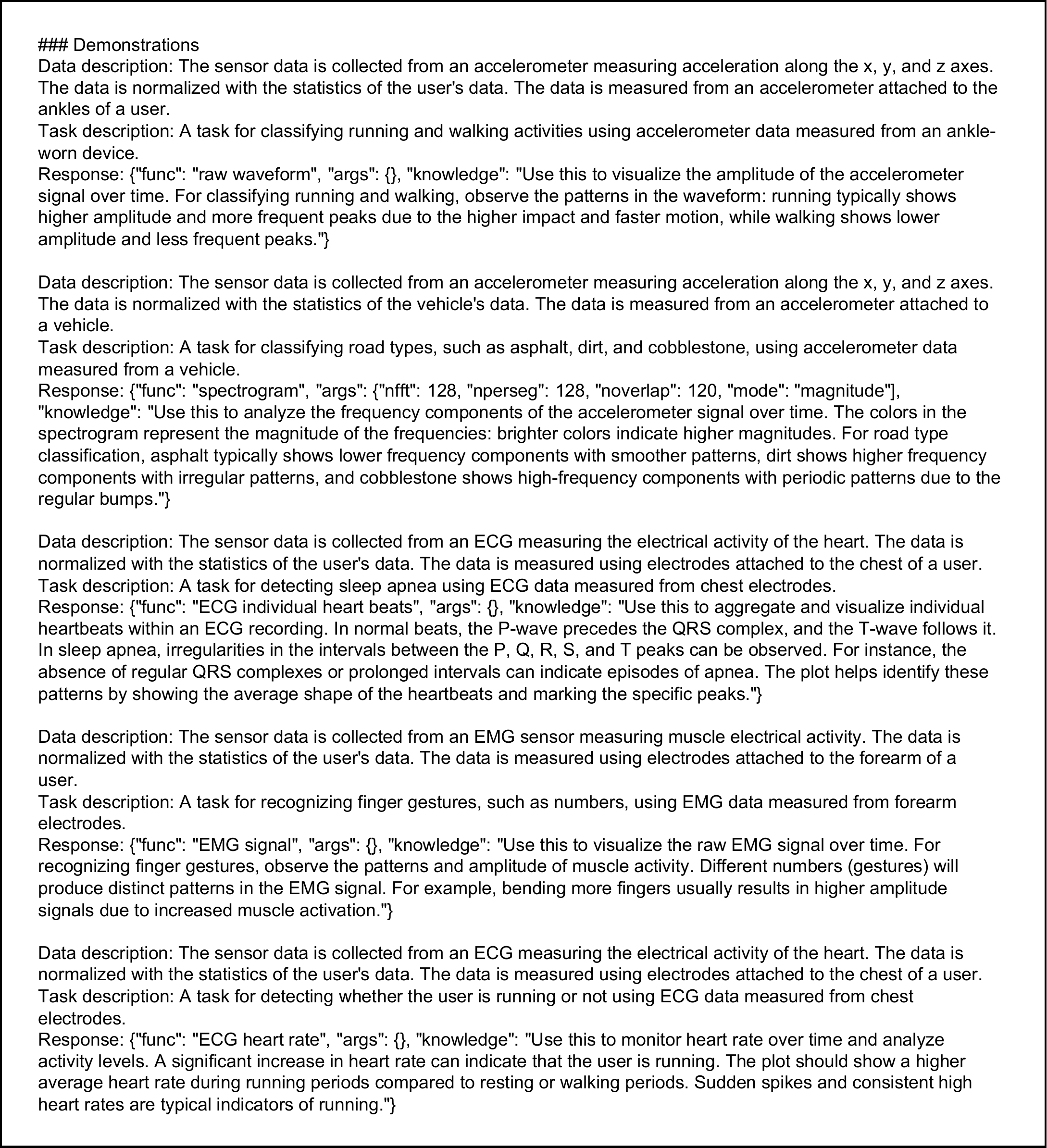}
    \caption{Demonstrations provided inside the visualization tool filtering prompt to enhance the response quality.}
    \label{fig:appendix_demonstrations}
\end{figure*}

\begin{figure*}
    \centering
    \includegraphics[width=\textwidth]{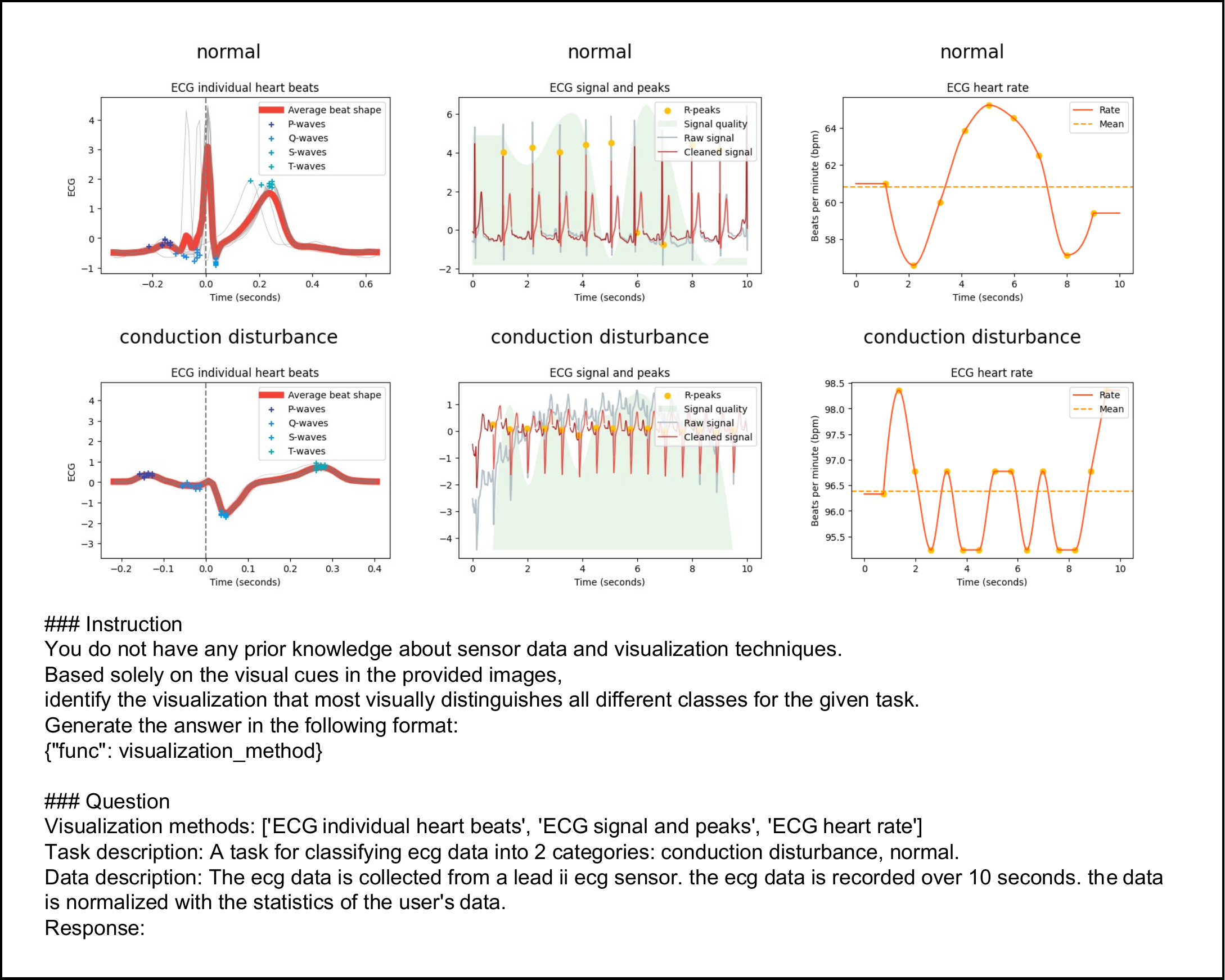}
    \caption{An example prompt from our visualization generator for visualization selection in the PTB-XL (CD) task.}
    \label{fig:appendix_visualization_selection}
\end{figure*}

We present examples of prompts used in our experiments. Figure~\ref{fig:appendix_txtprompt_hhar} and Figure~\ref{fig:appendix_txtprompt_ptbxl} illustrate two text-only prompt examples derived from the HHAR and PTB-XL (CD) datasets; in these examples, sensor data is truncated after a certain point to conserve space, though the format remains consistent with varying values. Figure~\ref{fig:appendix_visprompt_hhar} and Figure~\ref{fig:appendix_visprompt_ptbxl} displays the visual prompts created for the same datasets, HHAR and PTB-XL (CD). Figure~\ref{fig:appendix_tool_filtering} details the prompt for our visualization tool filtering specific to the PTB-XL (CD) task, with demonstrations omitted and presented separately in Figure~\ref{fig:appendix_demonstrations}. Lastly, Figure~\ref{fig:appendix_visualization_selection} showcases the visualization selection prompt for the PTB-XL (CD) dataset.

\end{document}